\documentclass[10pt,twocolumn,letterpaper]{article}

\usepackage[pagenumbers]{cvpr} 

\usepackage{times}
\usepackage{epsfig}
\usepackage{graphicx} 
\usepackage{amsmath}
\usepackage{amssymb}
\usepackage{floatrow}

\usepackage{xcolor}
\usepackage{subcaption}
\usepackage{color}
\usepackage{bbold}

\definecolor{old}{rgb}{0.7,0.7,0.7}
\newcommand{\old}[1]{\textcolor{old}{}} 

\definecolor{yannick}{rgb}{0.,0.5,0.9}

\definecolor{ben}{rgb}{0.9,0.,0.5}

\definecolor{barnabe}{rgb}{0.,0.9,0.5}

\definecolor{sgm}{rgb}{0.,0.,0.9}

\definecolor{jifei}{rgb}{0.,0.25,0.25}

\definecolor{todo}{rgb}{1.0, 0., 0.}

\usepackage{adjustbox}
\usepackage{chngpage}
\usepackage{multirow}
\usepackage{amssymb}
\usepackage{graphicx}
\usepackage[shortlabels]{enumitem}
\usepackage{array}
\usepackage{tabularx,booktabs}
\usepackage[percent]{overpic}
\usepackage{stackengine,graphicx}

\usepackage{wrapfig} 
\usepackage{stackengine}
\usepackage{lipsum}
\usepackage{floatrow}

\def\DA{\smash{\bclap{\scalebox{1.6}{$\downarrow$}}}}

\newfloatcommand{capbtabbox}{table}[][\FBwidth]
\usepackage[pagebackref=true,breaklinks=true,colorlinks,bookmarks=false]{hyperref}
\usepackage[labeled,resetlabels]{multibib} 
\newcites{S}{References} 

\usepackage[capitalize]{cleveref}
\crefname{section}{Sec.}{Secs.}
\Crefname{section}{Section}{Sections}
\Crefname{table}{Table}{Tables}
\crefname{table}{Tab.}{Tabs.}

\def\cvprPaperID{3717} 
\def\confName{CVPR}
\def\confYear{2022}

\begin{document}

\title{CroMo: Cross-Modal Learning for Monocular Depth Estimation} 

\author{
\hspace{-12pt}
Yannick Verdi\'e$^{1}$, 
Jifei Song$^{1}$, 
Barnab\'e Mas$^{1,2}$, 
Benjamin Busam$^{1,3}$, 
Ale\v{s} Leonardis$^{1}$, 
Steven McDonagh$^{1}$\\
$^1$ Huawei Noah's Ark Lab\qquad
$^2$ \'Ecole Polytechnique\qquad
$^3$ Technical University of Munich\\
{\tt\small \{yannick.verdie,jifei.song,ales.leonardis,steven.mcdonagh\}@huawei.com}\\
\quad {\tt\small barnabe.mas@polytechnique.edu}
\quad {\tt\small b.busam@tum.de}
\vspace{-10pt}
}

\maketitle 

\begin{abstract}\noindent


Learning-based depth estimation has witnessed recent progress in multiple directions; from self-supervision using monocular video
to supervised methods offering highest accuracy. Complementary to supervision, further boosts to performance and robustness are gained by combining information from multiple signals. In this paper we systematically investigate key trade-offs associated with sensor and modality design choices as well as related model training strategies. Our study leads us to a new method, capable of connecting modality-specific advantages from polarisation, 
Time-of-Flight and structured-light inputs. We propose a novel pipeline capable of estimating depth from monocular polarisation for which we evaluate various training signals. 
The inversion of differentiable analytic models thereby connects scene geometry with polarisation and ToF signals and enables self-supervised and cross-modal learning.

In the absence of existing multimodal datasets, we examine our approach with a custom-made multi-modal camera rig and collect CroMo; the first dataset to consist of synchronized stereo polarisation, indirect ToF and structured-light depth, captured at video rates. Extensive experiments on challenging video scenes confirm both qualitative and quantitative pipeline advantages where we are able to outperform competitive monocular depth estimation methods.

\end{abstract}

\vspace*{-5mm} 
\vspace{-4pt}
\section{Introduction}
\vspace{-4pt}
\label{sec:intro}
\captionsetup[subfigure]{labelformat=empty}
\begin{figure}[h!]
\caption{Top row: polarisation input signal (Pol\onedot) visualised as Angle and Intensity. Additionally; Time-of-Flight Amplitude (i-ToF) and structured-light sensor co-modalities, exploitable during training. Bottom row: monocular depth estimation, using the Pol\onedot input. Uni-modal model training of p2d~\cite{blanchon2020p2d} and the monodepth2 architecture~\cite{godard2019digging}, compared with cross-modal training (Ours).
}
\label{fig:teaser}
\begin{adjustbox}{minipage=1.1\linewidth,scale=0.9}
  \centering
  \begin{adjustbox}{minipage=0.98\linewidth}
 \subfloat[\hfill]{\rotatebox{90}{
  \parbox{4.5em}{ \centering \footnotesize Training modalities}}}
  \hspace*{4pt}%
 \subfloat[\centering \footnotesize Pol. Angle ]{\includegraphics[width=0.22\linewidth]{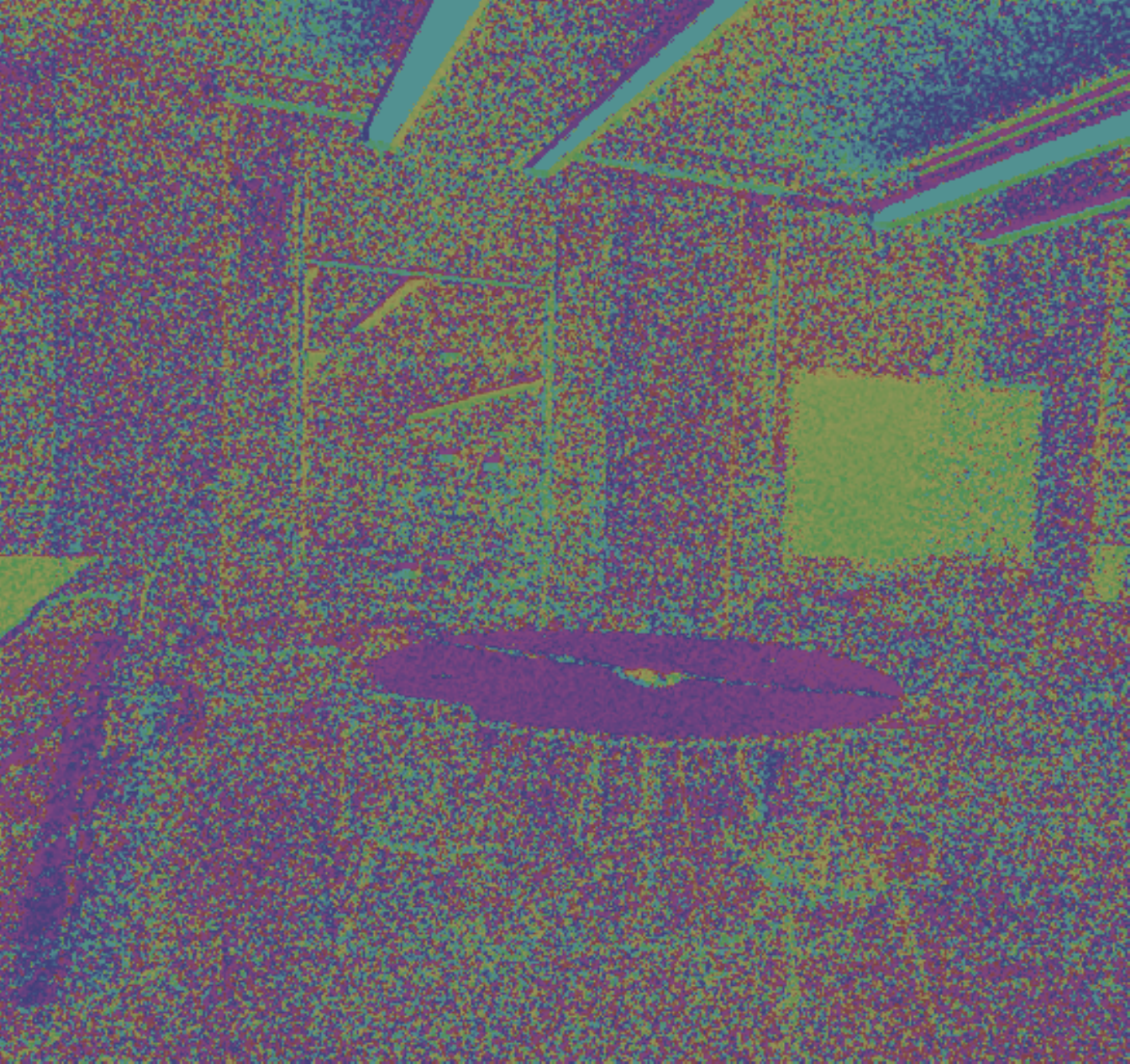}}
  \hspace*{\fill}%
 \subfloat[\centering \footnotesize Pol. Intensity]{
\includegraphics[width=0.22\linewidth]{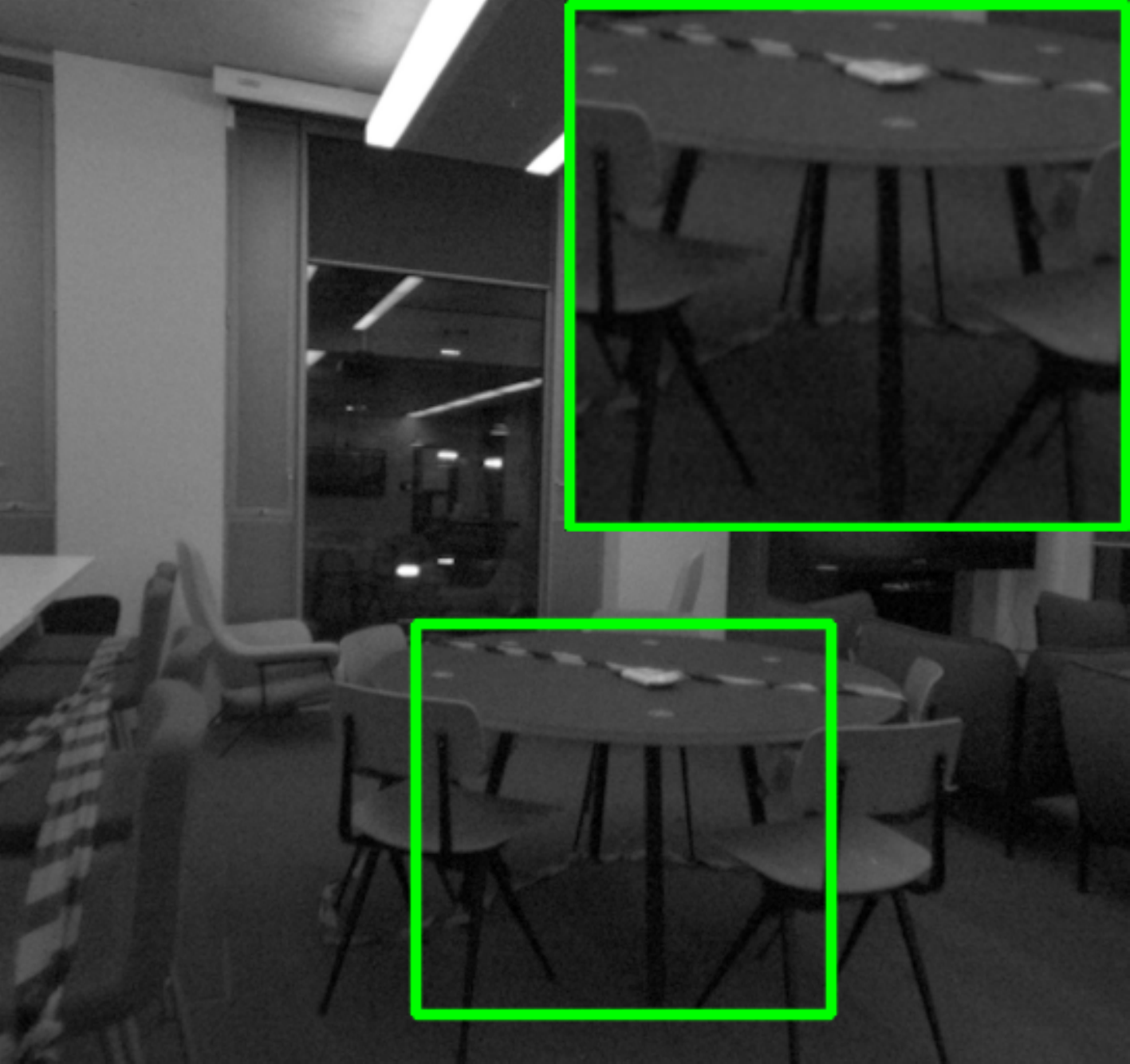}}
  \hspace*{\fill}%
 \subfloat[\centering \footnotesize i-ToF Amplitude]{
\includegraphics[width=0.22\linewidth]{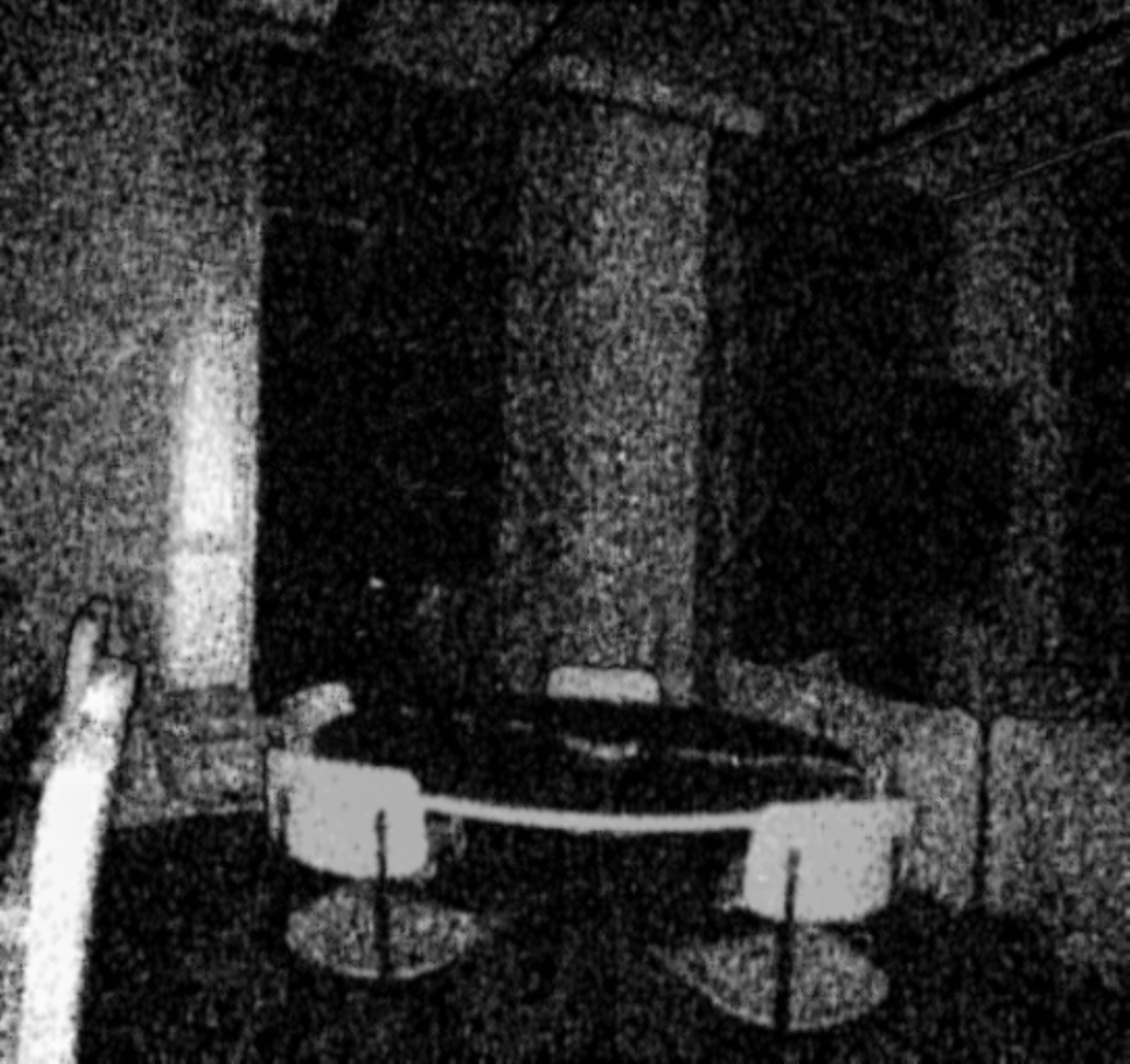}}
  \hspace*{\fill}%
 \subfloat[\centering \footnotesize Structured-light]{
\includegraphics[width=0.22\linewidth]{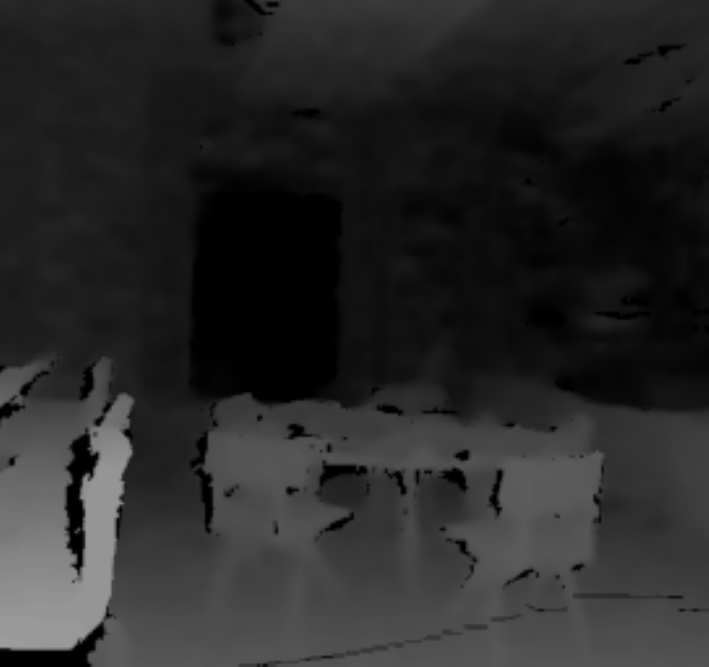}
}
\end{adjustbox}
 \begin{adjustbox}{minipage=1.1\linewidth,scale=0.9}
\subfloat[\hfill]{
\rotatebox{90}{
\parbox{5.5em}{ \centering \footnotesize Predictions and~GT}}}
  \hspace*{2pt}%
\subfloat[\centering \footnotesize P2d~\cite{blanchon2020p2d}]{
\includegraphics[width=0.22\linewidth]{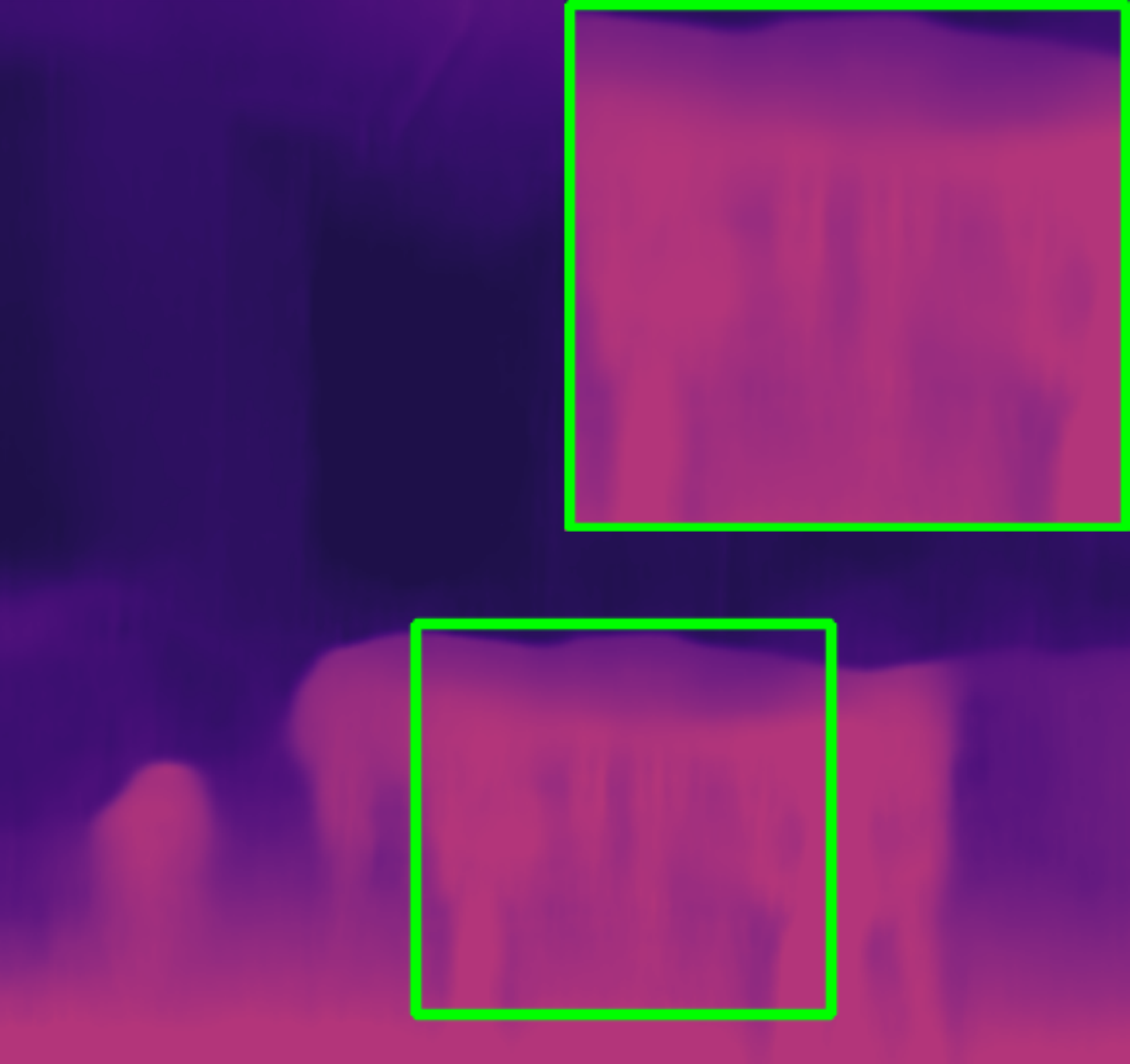}}
  \hspace*{\fill}%
\subfloat[\centering \footnotesize Monodepth2~\cite{godard2019digging}]{
\includegraphics[width=0.22\linewidth]{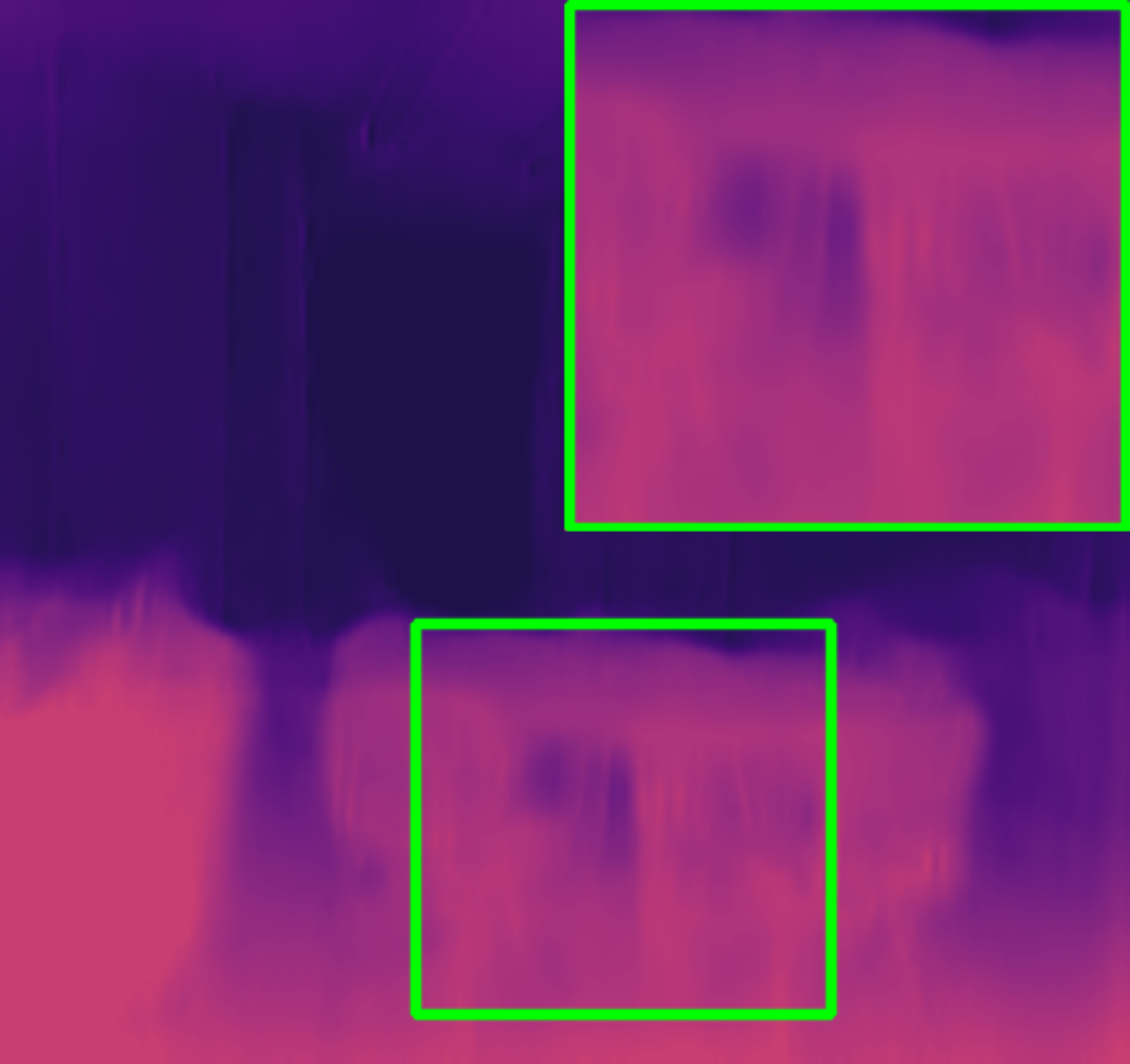}}
  \hspace*{\fill}%
\subfloat[\centering \footnotesize Ours]{
\includegraphics[width=0.22\linewidth]{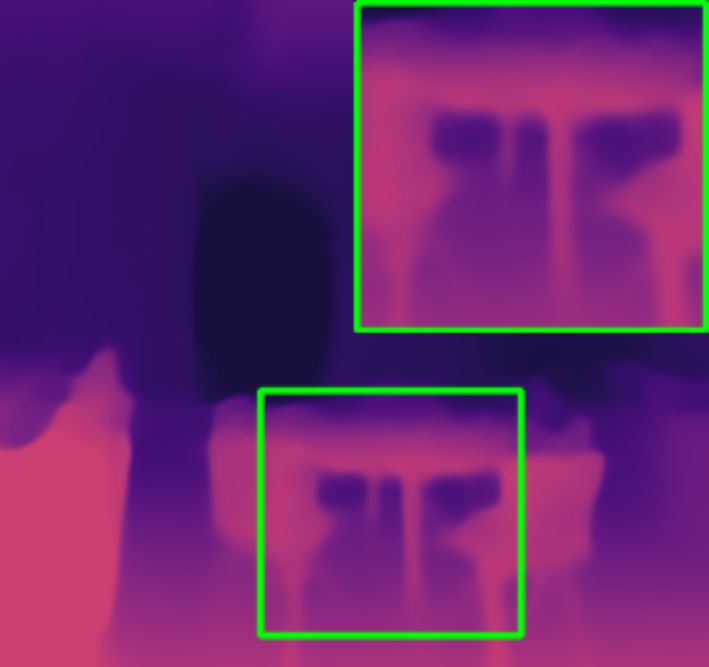}}
  \hspace*{\fill}%
\subfloat[\centering \footnotesize GT Label]{
\includegraphics[width=0.22\linewidth]{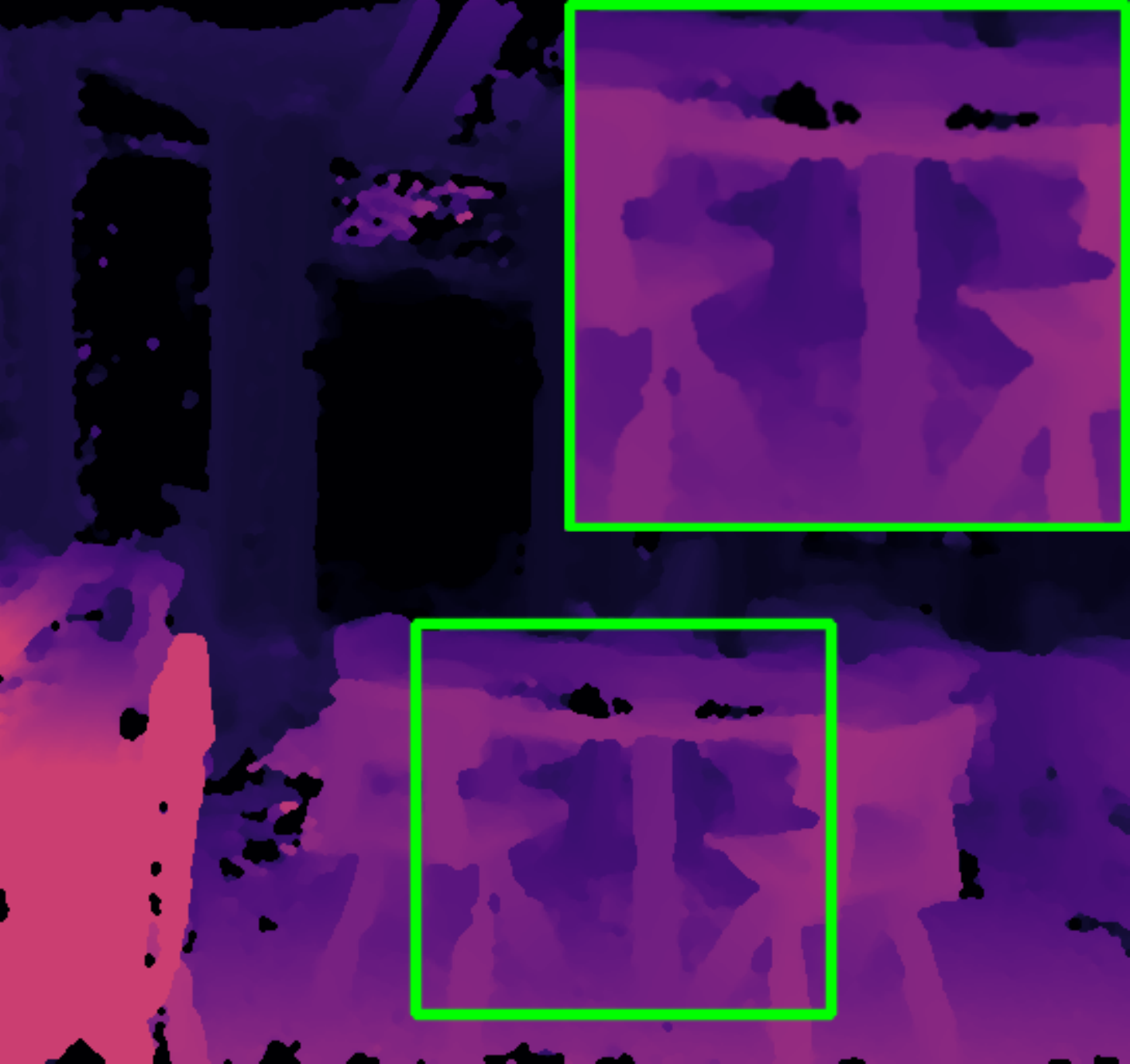}
}
\end{adjustbox}
\end{adjustbox}
\end{figure}
\captionsetup[subfigure]{labelformat=simple, labelsep=period}

Modern vision sensors are able to leverage a variety of light properties for optical sensing. Common RGB sensors, for instance, use colour filter arrays (CFA) 
over a pixel sensor grid to separate incoming radiation into specified wavebands. This allows a photosensor to detect wavelength-separated light intensity and enables the acquisition of familiar visible spectrum images. 
Wavelength is however only one property of light capable of providing information.

Light polarisation defines another property and describes the oscillation direction of an electromagnetic wave. 
While the majority of natural light sources (\eg the sun) emit unpolarised light, consisting of a mixture of oriented oscillations, surface reflection from non-metallic objects can linearly polarise the light. 
Such polarised light then contains surface structure information, retrievable using analytic physical models~\cite{atkinson2006recovery}. 
This information can be used to harness the depth cues offered by this light property. Polarimetric imagery is a \emph{passive} example for depth estimation. Passive sensors have acceptable resolution and dense depth however there exist well understood capture situations that prove challenging 
(\eg textureless surface regions).

However further known properties of light (\ie speed) provide yet more information. Indirect Time-of-Flight (i-ToF) cameras are \emph{active} light sensors and use a pulsed, near infrared light source to measure object and surface distance. 
Further active sensors use structured-light and these emit known infrared patterns and use stereoscopic imaging to measure the distance to the surface.
While i-ToF and structured-light cameras have clear advantages, such as the ability to function in low-light scenarios and good short range precision, they are susceptible to specular reflections, ambient light and range remains limited.

We argue that novel combination of \emph{active} and \emph{passive} light sensors 
offers new possibilities. We can exploit such a combination to take advantage of the discussed, modality-specific strengths and weaknesses. 
We observe that (1) differing visual modalities offer information cues about complimentary aspects of the world and (2) there exist clear trade-offs between the complexity of capture sensor setups and the resulting 
data diversity and quality, accessible for supervision signals. 
This motivates us to systematically investigate these considerations and provide insight into training data capture design decisions and the related pay-offs. Our study results in the proposal of a framework capable of exploiting available supervision signals and is tailored to benefit from the particular strengths of unique modalities. 

We instantiate our ideas by bringing together the physical understanding of Polarisation and i-ToF in a data driven fashion. In practice this affords an inference pipeline that estimates depth from a single polarisation image. We train a convolutional neural network (CNN), 
with cross-modal fusion using differentiable physical models. We establish a dataset comprising Ground Truth depth obtained via 
Multi-View Stereo (MVS) reconstruction~\cite{schoenberger2016mvs} that enjoys access to information rich, full video sequences. We carry out extensive experimental work to establish the efficacy of our proposed monocular depth estimation strategies. 
\newline
\noindent Our \textbf{contributions} can be summarised as:
\begin{enumerate}[noitemsep,topsep=0pt,nosep,leftmargin=*]
   
\item \textbf{Novel multi-modal method}.
   {We propose a multi-modal training approach that allows for monocular depth estimation from polarisation images. We propose (i) differentiable analytic formulae that define modal-specific loss terms, 
   (ii) cross-modal consistency joint-training towards improved real-world depth estimation from a single polarisation image, (iii) 
   architectural components that increase predicted depth sharpness (see Fig.~\ref{fig:teaser}).
   }
   
\item \textbf{CroMo dataset and training modalities study}.
   {We provide a systematic analysis of the benefits afforded when multiple image modalities are available at training time, for monocular depth estimation. Investigation and exposure of improvements are enabled by the unique Cross-Modality video dataset\footnote{Dataset is available at: \url{https://cromo-data.github.io/}}. Our multiple hardware-synchronized cameras capture, for the first time, stereo polarisation (Pol), indirect Time-of-Flight (i-ToF) and 
   structured-light images from active sensing. 
   }
   
\end{enumerate}

\noindent The remaining sections of the paper are thus organised: Sec.~\ref{sec:related} provides brief review of depth estimation with respect to relevant modalities and previous work considering multiple information signals. Sec.~\ref{sec:method} presents our model capable of 
monocular depth estimation from polarisation imagery and our cross-modal training procedure. In Sec.~\ref{sec:dataset} we introduce CroMo, our novel multi-modal dataset, Sec.~\ref{sec:experiments} reports experimental work validating our contributions and Sec.~\ref{sec:conclusion} provides discussion and future research issues.
\vspace{-4pt}
\section{Related Work}
\vspace{-4pt}
\label{sec:related}

To the best of our knowledge this is the first work to study end-to-end monocular depth inference, utilising cross-modal information from Time-of-Flight (i-ToF), active stereo and polarisation modalities during training. 
We briefly review the literature most closely related to the main components of our investigation and proposed framework. 

\vspace{-4pt}
\subsection{Monocular depth estimation}
\vspace{-4pt}
Estimating depth from a single image constitutes a hard, ill-posed problem. 
Pioneering work on supervised monocular depth estimation~\cite{mayer2018makes} used synthetic samples during training. Synthetic data was also previously used in conjunction with stereo network distillation~\cite{guo2018learning} for this task. To improve accuracy and convergence speed, \cite{fu2018deep} introduce a spatially-increasing discretisation. However, acquiring ground truth depth data remains a difficult task~\cite{geiger2012we}.

To overcome the difficulty of collecting accurate ground truth signal, multiple works~\cite{xie2016deep3d,garg2016unsupervised} investigate a consistency loss by leveraging stereo imagery during training, towards self-supervision. 
While being undoubtedly path-breaking, the initial methods suffered from a non-differentiable sampling step. 
Godard \etal.~\cite{monodepth17} formulated a fully-differentiable pipeline with left-right consistency checks during training and have also explored the temporal components~\cite{godard2019digging}, even in challenging setups such as night scenes~\cite{spencer2020defeat}. These methods predict depth with RGB input, while we utilise polarisation images.
\noindent\textbf{Monocular Polarisation}
Previous work use monocular polarisation imagery to recover depth. One route to overcome Shape from Polarisation (SfP) ambiguities is to use orthographic camera models to express polarisation intensity in terms of depth~\cite{yu2017shape}. Atkinson \etal~\cite{atkinson2006recovery} compute depth without knowing the light direction through a non-linear optimization framework and yet assume fully diffuse surfaces. Linear systems have also been constructed for the task~\cite{smith2018height} by adding shape from shading equations. While theoretically interesting, the orthographic assumption has restricted their application to synthetic lab environments.



\noindent\textbf{Learning based Polarisation} 
Due to lack of reasonably-sized datasets, only a limited number of works focus on learning with polarisation. 
Ba \etal~\cite{ba2019physics} provide polarisation images together with a set of plausible inputs from a physical model to estimate surface normals.
The work of~\cite{kalra2020deep} apply polarisation for instance segmentation of transparent objects and~\cite{lei2020polarized} learn de-glaring of images with semi-transparent objects. Recently, Blanchon \etal~\cite{blanchon2020p2d} extended the work of~\cite{godard2019digging} with complementary polarimetric cues. In contrast to them, we invert a physical model to enable self-supervision through consistency cycles and additionally study the benefit of co-modal i-ToF information.

\noindent\textbf{Learning based i-ToF} 
i-ToF sensors acquire distance information by estimating the time required for an emitted light pulse to be reflected~\cite{zanuttigh2016time}. Sensors measure either the time (direct) or the phase (indirect) difference between emitted and received light. The modality enjoys high precision for short range distances~\cite{hansard2012time}, yet suffers from limited spatial resolution and noise~\cite{foix2011lock}, which constitute challenging factors for any learning-based approach. Obtaining reliable signals from specular surfaces is difficult and inherent Multi Path Interference (MPI), often manifests as noisy measurements and artifacts.
Synthetic training is also explored for raw i-ToF input data in end-to-end  fashion~\cite{su2018deep,guo2018tackling,agresti2018deep}. However, the ability to account for real world domain shifts is limited. In~\cite{Agresti_2019_CVPR} a GAN is employed towards addressing such domain adaption issues on a limited dataset. 

\noindent\textbf{i-ToF depth improvement}
MPI can be considered a critical issue and error mitigation has been the focus of a body of work~\cite{whyte2014review}.
Two-path approximations~\cite{godbaz2012closed} have been used within optimization schemes~\cite{dorrington2011separating,kirmani2013spumic} and multiple frequencies are used to constrain the problem~\cite{freedman2014sra}. 
Kadambi \etal.~\cite{kadambi2013coded} propose a hardware solution to address scenes with translucent objects and a number of scholars incorporate light transport information to correct for MPI~\cite{o2014temporal,gupta2015phasor,naik2015light,achar2017epipolar}.

\vspace{-4pt}
\subsection{Depth with multiple sensors}
\vspace{-4pt}
\noindent\textbf{Depth completion} has been carried out via combining multiple input modalities, for example, a sparse but accurate LiDAR signal in combination with RGB~\cite{uhrig:2017}. It is difficult to address sparse signals with CNNs~\cite{ma:2018} and LiDAR sensors can produce problematic artifacts resulting in unreliable Ground Truth depth estimates~\cite{lopez2020project}.
One strategy towards removing dependence on this form of supervision are self-supervision cues however these fall behind supervised pipelines in terms of accuracy~\cite{ma:2019}.

\noindent\textbf{i-ToF and \textit{x}}
Confidence-based combination of i-ToF depth and classical RGB stereo is explored with the network architecture of~\cite{agresti:2017} and a semi-supervised approach for this combination is explored by~\cite{pu2018sdf} in a generic framework. While these approaches improve upon the individual depth estimates, they rely on a late fusion paradigm. Son~\etal.~\cite{son2016learning} use a robotic arm to collect $540$ real training images of short range scenes with structured light ground truth.

By inserting micro linear polarizers~\cite{nordin1999micropolarizer} in front of a photo-receiver chip, Yoshida \etal.~\cite{yoshida2018improving} build an i-ToF sensor capable of acquiring both i-ToF depth and polarisation scene cues. Combination of both the absolute depth (i-ToF) and relative shape (polarisation cues) allowed reconstruction of depth for specular surfaces. 
While this pipeline requires i-ToF and polarisation input to solve an optimization problem, we alternatively explore cross-modal self-supervised training and single image inference.


\noindent\textbf{Depth from multi-view Polarisation}
Another route to predict depth is the use of more than one polarisation image~\cite{atkinson2005multi} which enables methods based on physical models.
An RGB+Polarisation pair can provide sharp depth maps with stereo vision~\cite{zhu2019depth}.
Other methods~\cite{cui2017polarimetric} use more than two polarisation images.
Despite the sharpness of the results, the difficulty to acquire multi-view polarisation images is still a major hurdle.
Atkinson \etal~\cite{atkinson2017polarisation}  combine polarisation methods with photometric stereo. Two images of a scene, from an identical view point yet with different light exposures, are leveraged.
An extension dealing with mixed reflectivity is established via a combined photometric-polarisation linear system in~\cite{logothetis2019differential} and Garcia \etal~\cite{garcia2015surface} solve for polarisation normals using circularly polarised light.
Traditional multi-view methods also benefit from polarisation. Miyazaki~\etal~\cite{miyazaki2016surface} recover surfaces of black objects using polarisation physics and space carving.

\noindent\textbf{Depth refinement with Polarisation}
Consumer depth estimation tools progress significantly in recent years however their predictions are noisy and lack details. Using polarisation cues, ~\cite{kadambi2017depth} enhance sharper depth maps from RGBD cameras by differentiating their depth maps to resolve polarisation ambiguities and perform mutual optimization.



Despite clear improvements in monocular depth estimation methods, their performance remains bounded by the chosen modality hence calling for multi-modal depth estimation. Our method alleviates this problem with a learning based approach. During training we leverage complementary modalities such that our model can compensate the drawbacks of the single modality used at inference time. 
\vspace{-4pt}
\section{Method}
\vspace{-4pt}
\label{sec:method}
\begin{figure*}
\vspace{-15pt}
\begin{adjustbox}{minipage=\linewidth,scale=0.9}
\begin{subfigure}[b]{0.42\textwidth}
 \centering
\includegraphics[page=1, trim = 20 300 10 280, clip,width=0.95\linewidth]{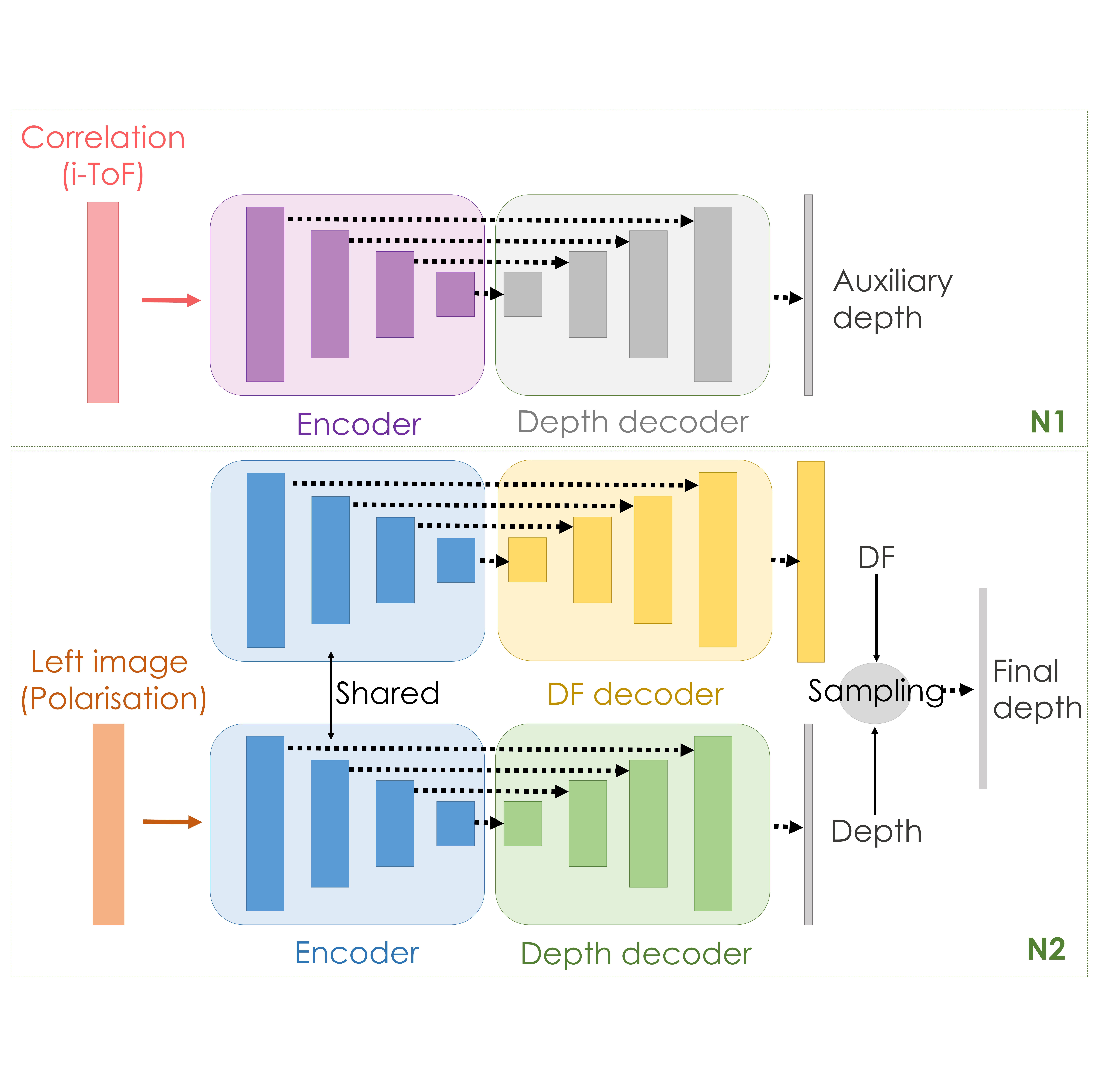}
\caption{Our network architecture.}
\label{fig:model}
\end{subfigure}
\qquad
\begin{subfigure}[b]{0.55\textwidth}
 \centering
\includegraphics[page=4, trim = 20 750 10 200, clip,width=0.95\linewidth]{loss.pdf}
\caption{Our training procedure and introduced losses.}
\label{fig:loss}
\end{subfigure}
\caption{Our full model with modality specific losses $\mathcal{L}_{\text{corr}}$, $\mathcal{L}_{\text{corr} \rightarrow \text{pol}}$ and $\mathcal{L}_{\text{stereo}}$ (see Sec.~\ref{sec:arch} for further details).}
\end{adjustbox}
\end{figure*}
Our multi-modal monocular depth investigation leads to a new model architecture that accounts explicitly for prediction blur and introduces two novel analytic losses. We discuss these components in the following sections.
\vspace{-4pt}
\subsection{Architecture}
\vspace{-4pt}
\label{sec:arch}
Our architecture employs multiple encoder-decoder networks illustrated in Fig.~\ref{fig:model}. We observe that monocular depth estimation methods often incur blurry image predictions and we address this problem by introducing architectural components that account for prediction blur. Firstly convolutions in our encoders are coupled with gated convolution. Our network then composes a traditional U-Net~\cite{ronneberger:2015} with skip connections and the gated convolutions~\cite{yu2019freeform}. The encoder utilises a ResNet~\cite{he:2016} style block, while the decoder is a cascade of convolutions with layer resizing. 

\begin{figure}[h!]
  \centering
    \includegraphics[width=0.75\textwidth]{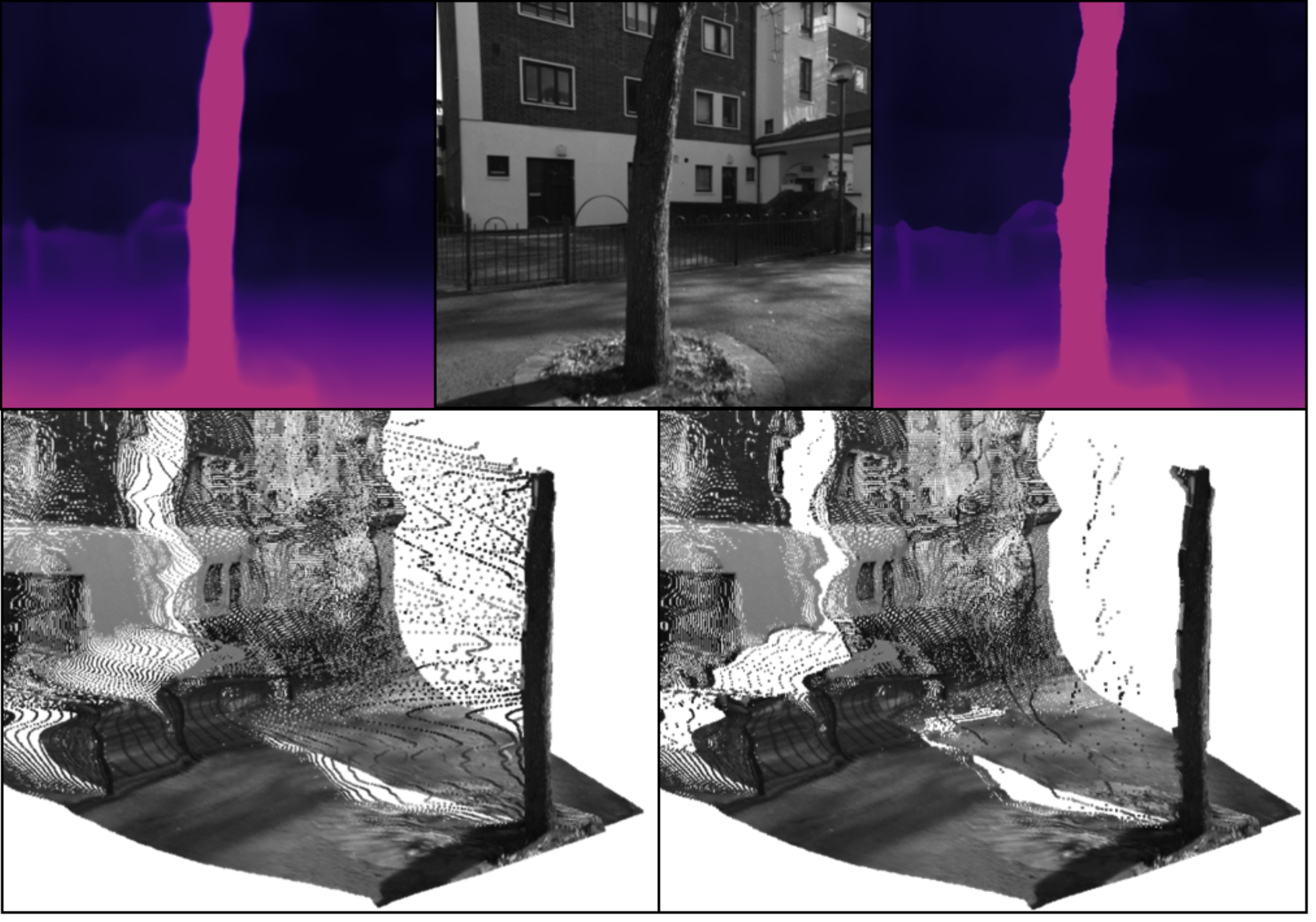}
  \caption{Effect of the \emph{Displacement Fields} (DF). Left top (bottom):~predicted depth (point cloud) without DF. Center top: polarisation intensity. Right top (bottom):~predicted depth (point cloud) with DF. Flying pixels, visible in 3D, are 
  clearly reduced.} 
  \label{fig:flow}
\end{figure}

Secondly, drawing on the fact that Displacement Fields (DF) can be utilised to aid sharpness~\cite{ramamonjisoa2020dispnet}, we estimate a DF using a self-supervised sharpening decoder. Depth pixels with strong local disparity have values redefined to mirror a nearest neighbour that does not exhibit strong local disparity. Groundtruth (GT) displacement fields can thus be defined for each predicted depth during training (``on-the-fly''), guiding our dedicated displacement field prediction. We inspect predicted depth with and without our DF strategy and observe significant improved sharpness, most evident when employing 3D visualisations (Fig.~\ref{fig:flow}).

\vspace{-4pt}
\subsection{Loss Formulation}
\vspace{-4pt}
\label{sec:loss}

Our study considers multiple modalities and various sensor configurations at training time. We explore several loss terms to exploit our unique setting (see Tab.~\ref{tab:sensor_results}). Loss terms in our training procedure are enabled through both coordinate frame projections (P1, P2) and analytic transforms (A1, A2) of individual network (N1, N2) outputs (see Figs.~\ref{fig:model}, ~\ref{fig:loss}). We firstly process input modalities individually using distinct networks. These ingest i-ToF correlation and left polarisation images respectively and output initial depth maps. We propose two analytic losses, derived from properties of i-ToF and polarisation, to train and link the networks. We train the i-ToF module without ground truth and also leverage the available multi-modal information through \emph{image recovery} via related analytical formulae (A1, A2). Strategically similar to previous work~\cite{Dai2021}, at inference time we require only a single modality (in our case polarisation), 
and can discard network N1 completely.

Terms $\mathcal{L}_{\text{corr} \rightarrow \text{pol}}$ and $\mathcal{L}_{\text{corr}}$ evaluate discrepancies between each input image and respective \emph{recovered} images, obtained using auxiliary and final depth maps (Fig.~\ref{fig:loss}). Our individual branches share information through the loss term  $\mathcal{L}_{\text{corr} \rightarrow \text{pol}}$. Explicitly, we recover a polarisation image from an \emph{auxiliary} depth map and then project this, using projection P1, to the polarisation sensor frame of reference via the final depth map $D_{\text{pol}}$. Finally our third loss term  $\mathcal{L}_{\text{stereo}}$ is used to train the polarisation network (N2) by comparing the right polarisation image, projected using the predicted depth $D_{\text{pol}}$, with the left polarisation image. We next provide details of our analytical formulae for \emph{image recovery} and the loss terms that enable our training procedure.\\


\noindent\textbf{Depth to polarisation (A2)}
Polarisation cameras capture polarised intensity along directions $\varphi_{pol}$. The measured intensity is given by~\cite{zhu2019depth}
\begin{equation}
\begin{split}
i_{\varphi_{pol}} &= i_{un} \cdot \ (1+\rho \  \cos(2\varphi_{pol}-2\phi)) \\
\hfill \text{ with } \varphi_{pol} &\in  \Big\{0,\frac{\pi}{4},\frac{\pi}{2},\frac{3\pi}{4}\Big\}
\end{split}
\label{eq:Polarisation fundamental} 
\end{equation}
where $\varphi_{pol}$ is the polariser angle, $i_{un}$ is the intensity of unpolarised light, $\rho$ is the degree of linear polarisation and $\phi$ is the Angle of Polarisation (AoP). The polarisation parameters $\rho \in \{ \rho_{s}, \rho_{d}\}$ and $\phi  \in \{ \phi_{s}, \phi_{d}\}$ depend on local reflection type, 
either \textit{diffuse} ($d$) or \textit{specular} ($s$) as follows:
\begin{equation}
    \left\{
    \begin{array}{l}
        \rho_{s} =  \frac
            {2\sin^{2}(\theta)\cos(\theta)\sqrt{\eta^{2}-\sin^{2}(\theta)}}
            {\eta^{2}-\sin^{2}(\theta)- \eta^{2}\sin^{2}(\theta) +2\sin^{4}(\theta)}
            \\
            \\
        \rho_{d} = \frac
            {(\eta-1/\eta)^{2}\sin^{2}(\theta)}
            {2+2\eta^{2}-(\eta+1/\eta)^{2}\sin^{2}(\theta)+4\cos(\theta)\sqrt{\eta^{2}-\sin^{2}(\theta)}}
    \end{array}
    \right.
\label{eq:rho}
\end{equation}
with $\theta \in [0, \pi/2]$ the viewing angle and $\eta$ the object refractive index, typically $1.5$, and
\begin{equation}
    \left\{
    \begin{array}{lll}
        \phi_{d} &= \alpha \ [\pi] &\text{if the pixel is diffuse}\\
        \phi_{s} &= \alpha + \frac{\pi}{2} \ [\pi] &\text{if the pixel is specular}
    \end{array}
    \right.
\label{eq:Polarisation phase}
\end{equation}
The $\pi\text{-ambiguity}$ is denoted as $[\pi]$ in \eqref{eq:Polarisation phase}, and $\alpha$ denotes the azimuth angle of the surface normal $\textbf{n}$.
Azimuth angle $\alpha$ and viewing angle $\theta$ are obtained as
\begin{equation}
    \cos(\theta) =\frac{\textbf{n} \cdot \textbf{v}}{\|\textbf{n}\|\|\textbf{v}\|}
    \quad\text{and}\quad 
    \tan(\alpha) = \frac{n_y}{n_x},
\label{eq:viewing angle}
\end{equation}

\noindent with $\textbf{v}$ the viewing vector defined as the vector pointing toward the camera center from the 3D point $P(x,y)$ corresponding to pixel $(x,y)$ with depth $d(x,y)$ and 
 $\textbf{n}$ the outward pointing normal vector, defined as the cross product of the partial derivatives with respect to $x$ and $y$~\cite{zhu2019depth}:
 \resizebox{1.02\linewidth}{!}{
 \begin{minipage}{1.07\linewidth}
  \vspace{-8pt}
\begin{equation}
\textbf{n}=\begin{bmatrix}
-f_y \partial_{x}d(x,y)\\
-f_x \partial_{y}d(x,y)\\
(x-c_x)\partial_{x}d(x,y) + (y-c_y)\partial_{y}d(x,y) + d(x,y)
\end{bmatrix}
\label{eq:viewing vector}
\end{equation} 
\end{minipage} 
  \vspace{2pt}
}
with $f_x, f_y, c_x, c_y$ the camera intrinsics.

Hence, from a given depth map 
$d$, one can compute the azimuth angle $\alpha$ and the viewing angle $\theta$ using \eqref{eq:viewing angle} and \eqref{eq:viewing vector}, followed by the polarisation parameters $\rho$ and $\phi$ with \eqref{eq:rho} and \eqref{eq:Polarisation phase}. The polarisation images for diffuse and specular surfaces  $\widehat{I}_{\text{pol}}^{\text{diffuse}}$ and $\widehat{I}_{\text{pol}}^{\text{specular}}$ are finally recovered using the calculated polarisation parameters with \eqref{eq:Polarisation fundamental}.\\

\noindent\textbf{Depth to correlation (A1)}
Indirect ToF measures the correlation between a known emitted signal and the received signal. The emitted signal at frequency $f_{\mathrm{M}}$ is a sinusoid:
\begin{equation}
g(t)=2\cos \left(2{\pi}f_{\mathrm{M}} t\right)+1
\end{equation}
and the signal, reflected by the scene, is of the form~\cite{Horaud_2016} 
\begin{equation}
f(t)=\alpha \cos \left(2{\pi}f_{\mathbf{M}} t+2{\pi}f_{\mathbf{M}} \tau\right)+\beta
\end{equation}
where the $\tau$ is the time delay between the emitted signal $g(t)$ and the reflected signal $f(t)$.
The i-ToF measurement $c(x)$ is the correlation between the two signals:
\vspace{-5pt}
\begin{equation}
\begin{split}
c(x) &=
\lim_{T\rightarrow\infty}\frac{1}{T}\int_{-\frac{T}{2}}^{{\frac{T}{2}}}f(t)g(t-x)\,dt\\
&= \alpha \cos \left(2{\pi}f_{\mathrm{M}}x+2{\pi}f_{\mathrm{M}}\tau\right)+\beta
\end{split}
\label{eq:multi_pol_sinusoid}
\vspace{-5pt}
\end{equation}

\noindent where we only consider the direct reflection signal and ignore the multipath interference (MPI) and sensor imperfections.
We are interested in the phase $\varphi$, proportional to the depth $d$ between the objects in the scene and the sensor:
\begin{equation}
\varphi = \left( 2\pi  f_{\mathrm{M}}\tau \right) [2\pi] = \left( d \cdot \frac{ 4  \pi  f_{\mathrm{M}}}{\textit{\textbf{c}}} \right) [2\pi]
\label{eq:phase_and_depth_corr}
\end{equation}
where $\textit{\textbf{c}}$ is the speed of light and $[2\pi]$ represents the $2\pi\text{-ambiguity}$.
Using the four bucket strategy~\cite{lange:2001} to sample $c(x)$ at four positions, where $2{\pi}f_{\mathrm{M}}x \in \{0, \frac{\pi}{2}, \pi, \frac{3\pi}{2} \}$, four measurements
$\{ c(x_{0}),c(x_{1}),c(x_{2}), c(x_{3}) \}$  can be obtained to recover the phase $\varphi$, the amplitude $\alpha$ and the intensity $\beta$.
\begin{align}
\tan(\varphi) &= \frac{c(x_{3}) - c(x_{1})}{c(x_{0}) - c(x_{2})}
\label{eq:phase_corr} \\
\alpha &=\frac{1}{2} \sqrt{(c(x_{3}) - c(x_{1}))^{2}+(c(x_{1}) - c(x_{0}))^{2}}
\label{eq:amplitude_corr} \\
\beta &= \frac{1}{4}\sum_{i=0}^{3} c(x_{i})
\label{eq:intensity_corr}
\end{align}
Hence, from a given depth $d$, one can compute the phase $\varphi$ using \eqref{eq:phase_and_depth_corr} and then reformulate the recovered i-ToF correlation using \eqref{eq:phase_corr}, \eqref{eq:amplitude_corr} and \eqref{eq:intensity_corr} in turn to form $\widehat{I}_{\text {corr}}$.



\noindent\textbf{Stereo loss $\mathcal{L}_{\text{stereo}}$}
This loss requires that left and right image pairs are accessible during training. While only the left image $I_{l}$ is fed to the network, the right image $I_{r}$ can guide the model towards generating valid depth, and vice versa. 
More formally, let $K_{l}$ and $K_{r}$ be camera matrices with intrinsic parameters for left and right images respectively, and $D$ a depth map on the left reference frame. Let $\text{T}_{\text{left} \rightarrow \text{right}}$ denote the transformation that moves 3D points from the left coordinate system to the right. An image coordinate transformed from left coordinate $p_l$ to the right image is
\begin{align}
    p_{\text{left} \rightarrow \text{right}} = K_r \cdot T_{\text{left}\rightarrow\text{right}} \cdot D\left(p_l\right) \cdot K_l^{-1} \cdot p_l
    \label{eqn:reprojection}
\end{align}
A backward differentiable warping~\cite{jaderberg:2015} is used to reproject an image onto the left view as
$I_{\text{right}{\rightarrow}\text{left}}$.

We form a stereo loss $\mathcal{L}_{\text {stereo}}$, and related mask loss $\mathcal{L}_{\text {mask}}$ similarly to~\cite{godard2019digging}, which aid network training and deal with occluded pixels respectively as
\begin{align}
 \mathcal{L}_{\text {stereo}} &= E_{\text{pe}}\left(I_{l}, I_{\text{right}\underset{D_{pol}}{\longrightarrow} \text{left}}\right)
 \label{eq:rep_error}\\
    \mathcal{L}_{\text {mask}} &= E_{\text{pe}}\left(I_{l}, I_{\text{right}\underset{D\infty}{\longrightarrow} \text{left}}\right)
    \label{eq:mask_error}
\end{align}
where the photometric error is similar to~\cite{godard2019digging}:

\scalebox{0.94}{\parbox{1.01\linewidth}{%
\begin{flalign}
    {E_{\text{pe}}}(I_x,I_y) &= \alpha\tfrac{1-\text{SSIM}(I_x, I_y)}{2} + (1-\alpha) \left \Vert I_x-I_y \right \Vert_{1}
\end{flalign}
}}


\noindent\textbf{Analytical losses $\mathcal{L}_{\text{corr}}$ and $\mathcal{L}_{\text{corr} \longrightarrow \text{pol}}$ }
Depth $D_{\text{corr}}$ is firstly inferred directly from i-ToF correlation input, and then two recovered images $\widehat{I}_{\text{corr}}$ and $\widehat{I}_{\text{pol}}$ are formed. Recovered images represent the `ideal' input for each modality, i-ToF and polarisation respectively, conditioned on the inferred depth. Since $\widehat{I}_{\text{pol}}$ is generated from $D_{\text{corr}}$, we reproject it using $D_{\text{pol}}$ to form a recovered final polarisation image $\widehat{I}_{\text{corr}\underset{D_{pol}}{\longrightarrow} \text{pol}}^{{{i}}}, {i \in \{\text{diffuse}, \text{specular}\}}$. 
In each case, discrepancies between the recovered image and the true input image provide a strong indication of the quality of the generated depth. We use this signal to guide the network. Formally 
\resizebox{.999\linewidth}{!}{
\begin{minipage}{1.03\linewidth}
\begin{align}
\mathcal{L}_{\text {corr}} &= E_{\text{pe}}\left(I_{\text {corr}},\widehat{I}_{\text {corr}}\right)\label{eq:error_corr}\\
\mathcal{L}_{\text {corr} \rightarrow \text{pol}} &=\min_{i \in \{\text{diffuse}, \text{specular}\}}  \Big\{ E_{\text{pe}}\left(I_{l},\widehat{I}_{\text{corr}\underset{D_{pol}}{\longrightarrow} \text{pol}}^{{{i}}}\right) \Big\}\label{eq:error_corr_to_pol}
\end{align}
\vspace{1pt}
\end{minipage}
}

\noindent where $I_{l}$ is the left polarisation image, $\widehat{I}_{\text{corr}\underset{D_{pol}}{\longrightarrow} \text{pol}}$ the recovered polarisation image aligned to $I_{l}$, $I_{\text {corr}}$ the i-ToF correlation input, and $\widehat{I}_{\text {corr}}$ the recovered correlation image.
We use a $\min$ operator for $\mathcal{L}_{\text {corr} \rightarrow \text{pol}}$ to lift the problem of classifying a pixel as \emph{diffuse} or \emph{specular} by computing both possibilities and letting the network select the best solution. 

Finally, following~\cite{watson-2019-depth-hints}, we use an additional loss $\mathcal{L}_{\text{struct}}$ in the objective function, derived from structured-light information (see appendix for further detail).


In summary, depending on the input modalities available at training time, we can add or remove the introduced losses $\mathcal{L}_{\text{corr}}$, $\mathcal{L}_{\text{corr} \rightarrow \text{pol}}$, $\mathcal{L}_{\text{stereo}}$ and $\mathcal{L}_{\text{struct}}$ as appropriate. We explicitly note that hyper parameter tuning for balancing of these loss terms is \emph{not} required, due to our formulation.


Our total loss $\mathcal{L}$ can then be formulated as:
\begin{equation}
    \mathcal{L}=\min_{i \in \{\text{mask}, \text{stereo}, \text {corr} \rightarrow \text{pol}, \text{struct}\}} \Big\{  \mathcal{L}_{\text {i}} \Big\} + \mathcal{L}_{\text{corr}} +
    \mathcal{L}_{\text{DF}}
    \label{eq::total_loss}
\end{equation}

\noindent where $\mathcal{L}_{\text{DF}}$ is the $\mathcal{L}_{\text {2}}$ norm between predicted and GT \textit{DF}. 

\vspace{-4pt}
\section{Data}
\vspace{-4pt}
\label{sec:dataset}
We next provide details on our custom camera rig (Sec.~\ref{sec:dataset:capture_rig}) and 
CroMo dataset (Sec.~\ref{sec:dataset:cromo}), comprising synchronised image sequences capturing multiple modalities, at video-rate across real-world indoor and outdoor scenes. 
\vspace{-4pt}
\subsection{Camera capture rig}
\vspace{-4pt}
\label{sec:dataset:capture_rig}
Our prototype custom-camera hardware rig is shown in Fig.~\ref{fig:camera_rig:a}. Our rig is constructed in order to capture synchronised data across multiple modalities including stereo polarisation, i-ToF correlation, structured-light depth and IMU. We rigidly mount two polarisation
cameras (Lucid PHX050S-QC) providing a left-right stereo pair, an i-ToF camera (Lucid HLS003S-001) 
operating at 25Mhz and a camera (RealSense D435i) for active IR stereo capture. All devices are connected with a hardware synchronisation wire resulting in time-aligned video capture at a frame rate of $10$fps. The left polarisation camera is the lead camera which generates the {\em genlock} signal and defines the world reference frame. Accurate synchronisation was validated using a flash-light torch and was further confirmed by the respectable quality observed from stereo Block Matching results~\cite{hirschmuller2005accurate}. The focus of all sensors is set to infinity, the aperture to maximum, and the exposure is manually fixed at the beginning of each capture sequence. The calibration on all four cameras' extrinsics, intrinsics, and distortion coefficients is done with a graph bundle-adjustment for improved multi-view calibration (see appendix for further details).
\vspace{-4pt}
\subsection{CroMo dataset}
\begin{figure*}[!t]
\begin{floatrow}
\ffigbox[\FBwidth][\FBheight]{%
\scalebox{0.80}{
\includegraphics[width=0.45\textwidth]{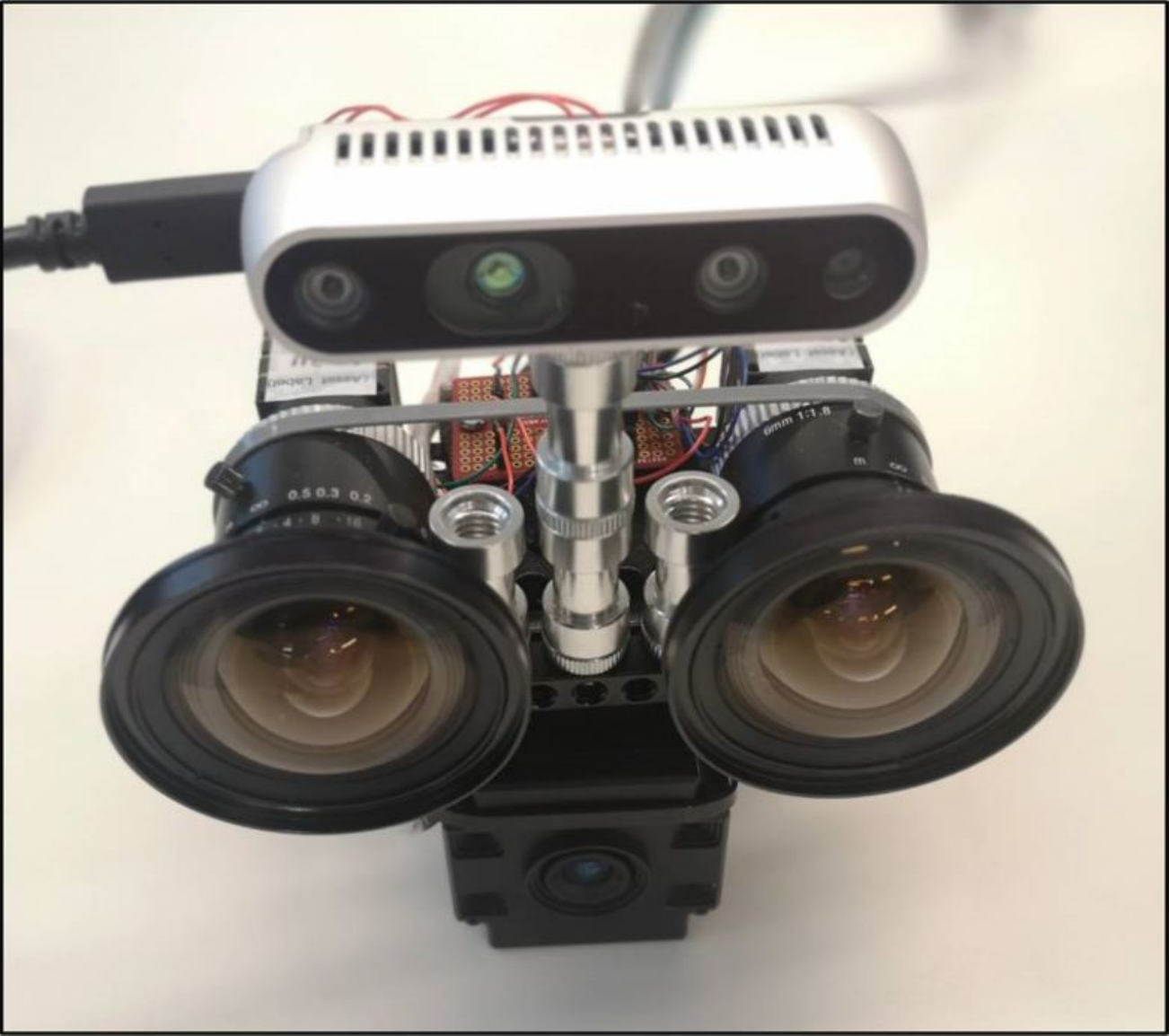}%
\put(-220,20){\parbox{7em}{ \centering Polarisation camera (Right)}}
\put(-70,30){\parbox{7em}{ \centering Polarisation camera (Left)}}
\put(-145,5){\parbox{7em}{ \centering i-ToF camera}}
\put(-190,185){\parbox{7em}{ \centering Structured-light camera}}
}
}
{%
\caption{Our multi-modal camera rig (see Sec.~\ref{sec:dataset:capture_rig}).}%
\label{fig:camera_rig:a}
}
\capbtabbox{%
\scalebox{0.65}{
\setlength\tabcolsep{2pt} 
\begin{tabularx}{0.675\textwidth}{r | c c c c c c c c c c } 
\toprule
  \small Dataset &
  \rotatebox[origin=c]{90}{\small RGB}          &
  \rotatebox[origin=c]{90}{\small Depth}        &
  \rotatebox[origin=c]{90}{\small IMU}          &
  \rotatebox[origin=c]{90}{\small i-ToF}        &
  \rotatebox[origin=c]{90}{\small Polarisation} &
  \rotatebox[origin=c]{90}{\small RAW}          &
  \rotatebox[origin=c]{90}{\small Real}         &
  \rotatebox[origin=c]{90}{\small Video}        &
  \rotatebox[origin=c]{90}{\small Available}    &
  \rotatebox[origin=c]{90}{\small Frames}       \\ 
\midrule
Sturm~\etal~\cite{sturm12iros}         & \checkmark & \checkmark   & \checkmark & -          & -          & -          & \checkmark & \checkmark & \checkmark & ${>}20k$   \\
Agresti~\etal~\cite{Agresti_2019_CVPR} & -          & (\checkmark) & -          & \checkmark & -          & -          & \checkmark & -          & \checkmark & $113$      \\
Guo~\etal~\cite{guo2018tackling}       & -          & \checkmark   & -          & \checkmark & -          & \checkmark & -          & -          & \checkmark & $2000$     \\
Zhu and Smith~\cite{zhu2019depth}      & (\checkmark) & -            & -          & -          & \checkmark & \checkmark & \checkmark & -          & \checkmark & $1$   \\
Qiu~\etal~\cite{qiu:2019a}             & \checkmark   & -            & -          & -          & \checkmark & \checkmark & \checkmark & -          & \checkmark & $40$  \\
Ba~\etal~\cite{ba:eccv2020}            & \checkmark   & (\checkmark) & -          & -          & \checkmark & \checkmark & \checkmark & -          & -          & $300$ \\
Kadambi~\etal~\cite{kadambi2017depth}  & \checkmark   & \checkmark   & -          & -          & \checkmark & \checkmark & \checkmark & -          & \checkmark & $1$   \\
\centering \textbf{ \stackunder{CroMo}{\DA}} & (\checkmark) & \checkmark & \checkmark & \checkmark & \checkmark & \checkmark & \checkmark & \checkmark & \checkmark & ${>}20k$    \\
\midrule
\multirow{2}{*}{\textbf{Scenes}} & &
\multicolumn{5}{c}{GT depth statistics (meters)}  
& 
& 
\multirow{2}{*}{\parbox[t]{0.8cm}{\centering valid ratio}} 
& 
\multirow{2}{*}{\parbox[t]{0.8cm}{\centering \# of seqs.}}
&
\multirow{2}{*}{\parbox[t]{0.8cm}{\centering \# of frames}} \\ 
 & 
 & \parbox[t]{0.8cm}{\centering \emph{mean}} 
 & \parbox[t]{0.8cm}{\centering\emph{var.} }
 & \parbox[t]{0.8cm}{\centering\emph{min} }
 & \parbox[t]{0.8cm}{\centering\emph{max} }
 & \parbox[t]{0.8cm}{\centering\emph{median}}
 &  & & &  \\
\bottomrule
\multicolumn{1}{r|}{Kitchen}         & & 3.3 & 3.6  & 0.3 & 15.7 & 2.9 &  & 0.95 & 3  & \multicolumn{1}{c}{2859}  \\ 
\multicolumn{1}{r|}{Station}         & & 4.9 & 14.8 & 0.3 & 18.9 & 3.6 &  & 0.86 & 11 & \multicolumn{1}{c}{7400}  \\ 
\multicolumn{1}{r|}{Facades}         & & 4.0 & 8.4  & 0.3 & 17.8 & 3.3 &  & 0.86 & 7  & \multicolumn{1}{c}{7228}  \\ 
\multicolumn{1}{r|}{Park}            & & 6.1 & 23.7 & 0.3 & 19.7 & 4.4 &  & 0.82 & 10 & \multicolumn{1}{c}{5551}  \\ 
\hline
\multicolumn{1}{r|}{Total}           & & 4.7 & 13.6 & 0.3 & 18.3 & 3.6 &  & 0.86 & 31 & \multicolumn{1}{c}{23038} \\ 
\end{tabularx}
}
}
{
\caption{CroMo comparison and dataset statistics.}%
\label{tab:dataset_comparison}%
}
\end{floatrow}
\end{figure*}

\vspace{-4pt}
\label{sec:dataset:cromo}
We collect a unique dataset comprising multi-modal captures such that each time point pertains to measurement of (1) \textbf{Polarisation}: raw stereo polarisation cameras produce $2448{\times}2048$~px stereo images. 
(2) \textbf{i-ToF}: 4 channel $640{\times}480$~px correlation images.
(3) \textbf{Depth}: a structured-light capture of the scene resulting in a $848{\times}480$~px depth image. 
In addition to the three main sensors, IMU information is recorded to further enable future research directions.
Our dataset consists of more than $20k$ frames, totalling ${>}80k$ images of indoor and outdoor sequences in challenging conditions, with no constraints on minimum or maximum scene range. We group these sequences into four different scenes which we name: \emph{Kitchen}, \emph{Station}, \emph{Park} and \emph{Facades}. Despite the multitude of senors, operating ranges are not unlimited and our data collection also does not cover all possible scenarios; we further discuss limitations in our appendix. We report statistics per captured scene in Tab.~\ref{tab:dataset_comparison} (lower).
These statistics characterise our scene captures and provide useful information, \eg, that the median scene depth differs greatly between indoor (\emph{Kitchen}) and outdoor (\emph{Station}, \emph{Park} and \emph{Facades}) scenes. This is a strong indicator for whether the i-ToF sensor will perform well. 
Tab.~\ref{tab:dataset_comparison} (upper) provides a comparison with other depth datasets showing that CroMo is the first publicly available, modality rich dataset containing a large quantity of image data.
\vspace{-4pt}
\section{Experiments}
\vspace{-4pt}
\label{sec:experiments}
Our experimental design evaluates (1) the effect of multiple modalities, accessible at training time, for monocular depth estimation and (2) the effect that changing network architecture has on depth quality, under consistent input signal.
Our capture setup allows us to employ a standard MVS approach~\cite{schoenberger2016mvs} on full temporal sequences of polarisation intensity frames (left-camera), to serve as ground-truth depth for our experimental work. This expensive offline optimisation leverages accordances amongst \emph{all} frames per sequence, affording high quality depth 
to evaluate our ideas.
\paragraph{Multi-modal training}
\vspace{-10pt}
We firstly evaluate combinations of training input signal by changing the number of sensors available to the model. We fix network encoder-decoder backbone components (\ie ResNet50, analogous to~\cite{godard2019digging}) and train models that leverage cues from a maximum of four sensors; left and right polarisation, i-ToF correlation and structured-light. We show predicted depth improvements, attainable by systematic addition of sensors, and quantify where best gains can be made.
The training signal components used for our monocular depth estimation experiments are as follow: \textbf{Temporal (M)} extracts information from video sequences (3 frames), \textbf{Stereo (S)} uses stereo images, 
\textbf{i-ToF (T)} 
leverages i-ToF correlations via our two interconnected depth branches (see Sec.~\ref{sec:loss}).
Finally, \textbf{Structured-light (L)} 
incorporates an additional mask into the objective function, derived from information provided by our structured-light sensor. The structured-light signal is utilised \emph{only when the mask improves the projection loss}. 
We explore alternative strategies to exploit the 
structured-light signal and discuss details on practical benefits (\eg convergence speed) 
in the appendix.

Introduced signal components 
define our set of training experiments. For example, \textbf{Stereo and Structure Light (SL)} train the model using self-supervised stereo \textbf{(S)} and structured-light \textbf{(L)} information. Experiments therefore use differing subsets of the introduced loss terms 
(see Tab.~\ref{tab:sensor_results}).

\begin{table*}[!t]
\scalebox{0.79}{
\begin{tabular}{l | c c | c c c | c c c } 
\toprule
{Models trained with \textbf{S}tereo \textbf{(S)} input} &
{\centering MP} &
{\centering GMACs} &
{\centering Sq Rel}  &
{\centering RMSE} & 
{\centering RMSE Log} & 
{\centering$\delta<1.25$}&
{\centering$\delta<1.25^2$}&
{\centering$\delta<1.25^3$}\\ 
\midrule
\small{ResNet18 architecture~\cite{godard2019digging}} & 14.36  & 20.17  
&1.7928	&2.1982	&0.3596	&0.5061	&0.7026	&0.8009\\
\small{ResNet50 architecture~\cite{godard2019digging}} & 32.55  & 39.62  
& 1.5037& 2.0642& 0.3383& 0.5324	& 0.7262& 0.8160 \\
\small{p2d~\cite{blanchon2020p2d} (ResNet50 - Stokes)} & 32.55 & 39.62 
&1.5938	&2.1291	&0.3884	&0.4565	&0.6632	&0.7775\\
\small{MiDaS architecture~\cite{Ranftl2020}} & 104.21 & 207.86 
&1.4021	&1.9985	&0.3252	&0.5409	&\textbf{0.7901}	&\textbf{0.8281}\\
\midrule
\small{Our architecture (\textit{\textbf{S}tereo (\textbf{S}) input})}
& 74.40  & 97.39  
& \textbf{1.3031} & \textbf{1.8889} & \textbf{0.3233} & \textbf{0.5533} & 0.7301 & 0.8213\\
\bottomrule
\end{tabular}
}
\caption{Architectural comparisons under consistent modality sensor input; \textbf{S}tereo \textbf{(S)}. 
Our proposed architecture improves quantitative results across the majority of metrics whilst remaining competitive in terms of compute and space requirements.}
\label{tab:arch_results}
\end{table*}
\begin{table*}[!t]
\scalebox{0.65}{
\begin{tabular}{ c | c | c c c c c | c c c | c c c } 
\toprule
{ \small{\shortstack{Image\\sensors}}} &
{\centering Training strategy} &
{\centering $\mathcal{L}_{\text {stereo}}$} &
{\centering $\mathcal{L}_{\text {DF}}$} &
{\centering $\mathcal{L}_{\text {corr}}$} &
{\centering $\mathcal{L}_{\text {corr} \longrightarrow \text {pol}}$} &
{\centering $\mathcal{L}_{\text {struct}}$} &
{\centering Sq Rel}  &
{\centering {RMSE}} & 
{\centering RMSE Log} & 
{\centering$\delta<1.25$}&
{\centering$\delta<1.25^2$}&
{\centering$\delta<1.25^3$} \tabularnewline
\midrule
\multirow{2}{*}{2}
 & \small{\textbf{S}tereo (\textbf{S}) \textit{w/o} DF sampling} & \checkmark & & & &
 & 1.5037& 2.0642& 0.3383& 0.5324	& 0.7262& 0.8160 \\

 & \small{\textbf{S}tereo (\textbf{S})} & \checkmark & \checkmark & & & 
 & 1.3031 & {1.8889} & {0.3233} & {0.5533} & 0.7301 & 0.8213

\tabularnewline
\midrule
\multirow{2}{*}{3}
 & \small{\textbf{S}tereo and i-\textbf{T}oF (\textbf{ST})} & \checkmark & \checkmark & \checkmark & \checkmark &
&1.2829	&1.8573	&0.3202	&0.5541	&0.7308	&0.9062  \tabularnewline
& \small{\textbf{S}tereo and Structured-\textbf{L}ight (\textbf{SL})} & \checkmark & \checkmark & & &\checkmark
&1.1233	&1.7510	&0.3168	&0.5529	&0.7370	&0.9251 \tabularnewline
\midrule
\multirow{2}{*}{4}
 & \small{\textbf{S}tereo, i-\textbf{T}oF, Structured-\textbf{L}ight \textbf{(STL)}} &  \checkmark &  \checkmark &  \checkmark & \checkmark & \checkmark
 & 1.0699&1.6070&0.2891&0.6512&0.7882&\textbf{0.9266} \tabularnewline 
 & \small{\textbf{STL}+Temporal \textbf{(STLM)}} &  \checkmark &  \checkmark &  \checkmark & \checkmark & \checkmark
&\textbf{1.0031}	&\textbf{1.4889}	&\textbf{0.2527}	&\textbf{0.7061}	&\textbf{0.8066}	&0.9246
\tabularnewline
\bottomrule
\end{tabular}
}
\caption{Model training strategies that differ in terms of available image sensor signals (utilised loss components). Sec.~\ref{sec:method} and~\ref{sec:dataset} provide details on loss function components and image sensors, respectively. In spite of having access to only a single, consistent modality during inference, the model benefits from visibility of additional training signals.}
\label{tab:sensor_results}
\end{table*}

\begin{figure}[t!]
\centering
\begin{adjustbox}{width=0.75\textwidth,center}
 \stackinset{l}{-12pt}{c}{5pt}{
 \rotatebox{90}{
 \parbox{5em}{ \centering \footnotesize GT Label}
 \parbox{5em}{ \centering \footnotesize STLM}
 \parbox{5em}{ \centering \footnotesize STL}
 \parbox{5em}{ \centering \footnotesize SL}
 \parbox{5em}{ \centering \footnotesize ST}
 \parbox{5em}{ \centering \footnotesize S}
 \parbox{5em}{ \centering \footnotesize Pol. Intensity}
 }}
  {%
  \stackinset{c}{}{b}{-15pt}{
 \parbox{5.5em}{ \centering \textit{Park}}
 \parbox{5.5em}{ \centering \textit{Facades}}
 \parbox{5.5em}{ \centering \textit{Kitchen}}
 \parbox{5.5em}{ \centering \textit{Station}}
 }
 {
\includegraphics[width=0.95\textwidth]{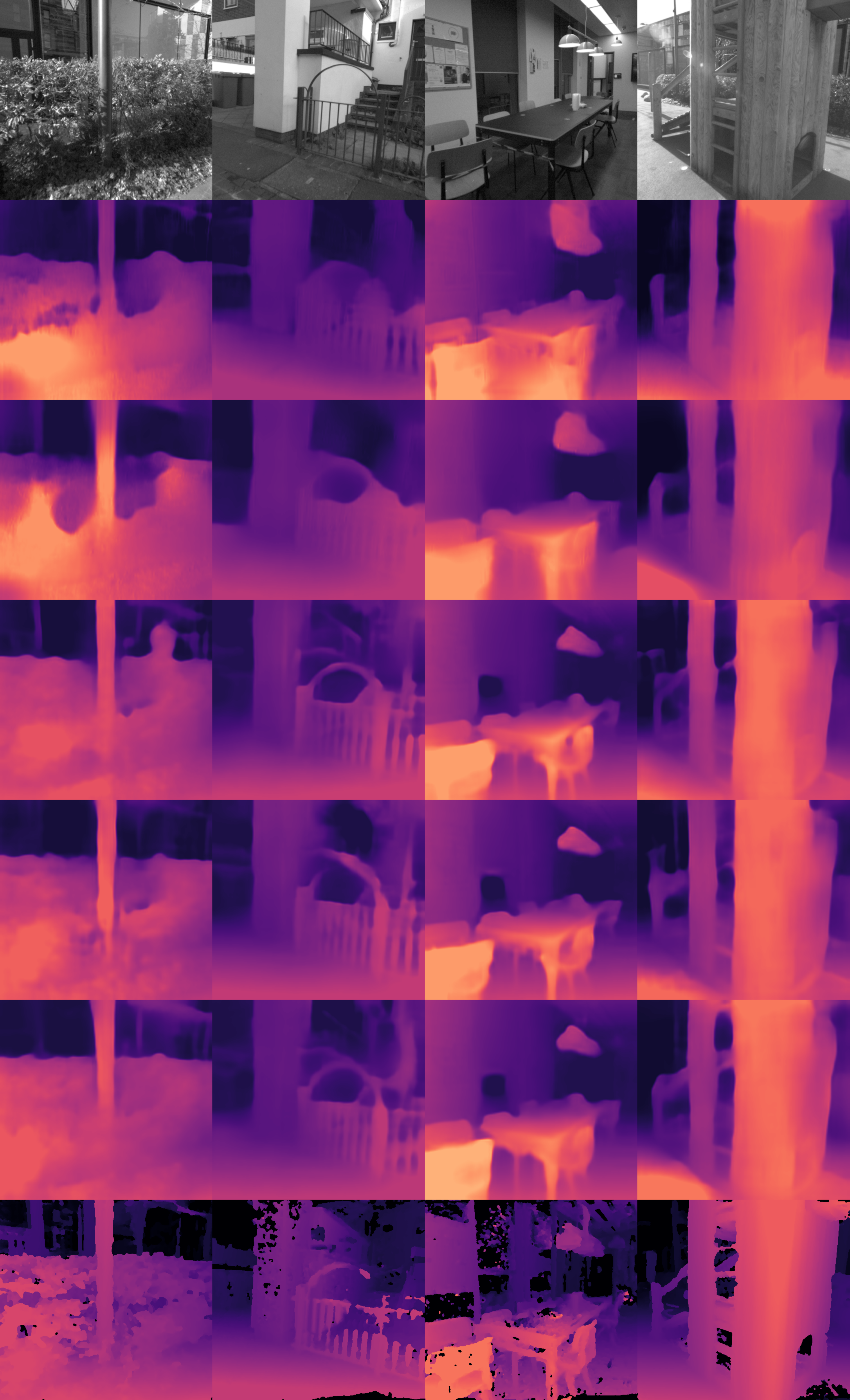}
}
}
\end{adjustbox}
\caption{Same ResNet50 architecture as in~\cite{godard2019digging} with different modalities: each new modality closes the gap with GT.}
\vspace{-4pt}
\label{fig:sensor_results}
\end{figure}

\noindent\textbf{Qualitative results} are shown in Fig.~\ref{fig:sensor_results}. Unsurprisingly, self-supervised stereo \textbf{(S)} is relatively blurry and struggles to capture fine details, such as the thin, metallic arch on the \textit{Facades} sample, or the furniture in the \textit{Kitchen}. 
Addition of i-ToF and structured-light modalities, exclusively at training time, 
result in (\textbf{ST}), (\textbf{SL}), (\textbf{STL}) and can be observed to improve respective depth quality. Finally, (\textbf{STLM}) adds our temporal modality 
and improves detail recovery (\eg metallic arch and fence). Qualitative results can be observed to corroborate our hypothesis; inclusion of additional modalities at training time 
afford the model multiple complementary depth cues that can qualitatively improve depth inference. 
Our experimental work highlights the nature of 
valuable 
investigation possible with our unique CroMo dataset.
\newline
\noindent\textbf{Quantitative results} are reported in Tab.~\ref{tab:sensor_results}. We follow~\cite{godard2019digging}, reporting standard evaluation metrics, with focus on the RMSE in our following experiments. Best performance is obtained when all sensors are used together (\textbf{STLM}) while self-supervision stereo (\textbf{S}) with only polarisation images performs worst. When additional modalities are added to self-supervision (\textbf{S}); \ie i-ToF (\textbf{ST}) or structure light (\textbf{SL}), performance improves in both cases, with larger gains come from the addition of the latter. We conjecture that structured light information helps more due to the nature of our dataset and current distribution of image content therein \ie ${\sim}85\%$ outdoor imagery, where i-ToF sensors are impaired by ambient light. 
Combining the i-ToF and structured-light sensors (\textbf{STL}), further improves. The best depth prediction utilises the temporal component (\textbf{STLM}). RMSE on Tab.~\ref{tab:sensor_results} displays a clear trend; the availability of additional sensor cues at training time improves monocular inference. 

\begin{figure}[t!]
\begin{adjustbox}{width=0.75\textwidth,center}
\stackinset{l}{-12pt}{l}{15pt}{
\rotatebox{90}{
\parbox{5em}{ \centering \footnotesize GT Label}
\parbox{5em}{ \centering \footnotesize Ours}
\parbox{5em}{ \centering \footnotesize MiDaS~\cite{Ranftl2020}}
\parbox{5em}{ \centering \footnotesize p2d~\cite{blanchon2020p2d} }
\parbox{5em}{ \centering \footnotesize ResNet50~\cite{godard2019digging} }
\parbox{5em}{ \centering \footnotesize ResNet18~\cite{godard2019digging}}
\parbox{5em}{ \centering \footnotesize Pol. Intensity}
}}
{%
\stackinset{c}{0pt}{b}{-15pt}{
\parbox{11em}{ \centering \textit{Station sample}}
\parbox{11em}{ \centering \textit{Park sample}}
}
{
\includegraphics[width=0.95\textwidth]{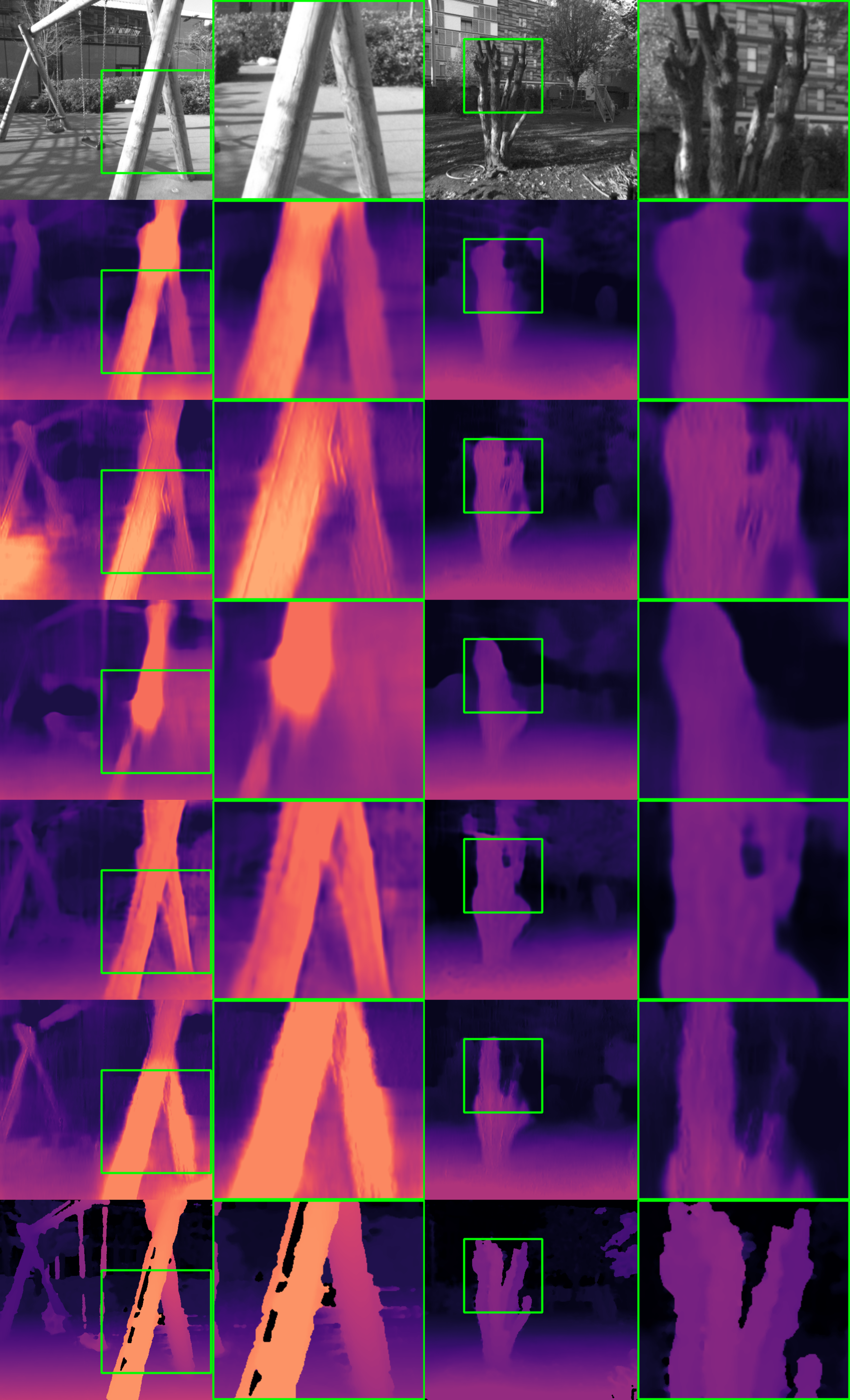}
}
}
\end{adjustbox}
\caption{Different architectures, same training strategy \textbf{S}tereo \textbf{(S)}: our new architecture produces the sharpest depth predictions.}
\vspace{-4pt}
\label{fig:arch_results}
\end{figure}
\vspace{-4pt}
\paragraph{Network architecture}
\vspace{-10pt}
We next investigate the effect of network architecture on monocular estimation performance. Of note, we highlight that employing a larger capacity network is not the only way to improve prediction performance. We use our self-supervised stereo \textbf{(S)}, \ie baseline-modality, training strategy for all experiments that follow in this section. 
This strategy provided \emph{weakest} performance in our previous investigation of training modality choice (Tab.~\ref{tab:sensor_results}). For this reason, we consider it an appropriate candidate with which to evaluate improvements afforded by changes to network architecture. We report \emph{millions-of-parameters} at inference (\textbf{MP}) and the \emph{giga-multiply-accumulates} per second (\textbf{GMACs}) in order to evaluate size and compute-cost per architecture. Architectures consist of the \textbf{ResNet18} U-Net used in~\cite{godard2019digging} and their supplementary material \textbf{ResNet50} variant, the \textbf{p2d} architecture~\cite{blanchon2020p2d} using \textbf{ResNet50} with a different data representation (Stokes), the \textbf{MiDaS}~\cite{Ranftl2020} architecture and \textbf{Ours} (see Sec.~\ref{sec:arch}, Fig.~\ref{fig:model}). 

\noindent\textbf{Qualitative results} are shown in Fig.~\ref{fig:arch_results}. It may be observed that the \textbf{ResNet18} architecture with smallest (\textbf{MP}) fails to obtain good background detail of the swing frame structure 
(\textit{Station} sample) or of the tree 
(\textit{Park} sample). The \textbf{ResNet50} variant slightly improves detail, especially with raw measurements instead of Stokes (\textbf{p2d}~\cite{blanchon2020p2d}). Even when increasing network capacity 
$c.$ three-fold with \textbf{MiDaS}, results are unsatisfying. Our proposed architecture (\textbf{Ours}) requires smaller capacity and 
computation for a sharper reconstruction of 
the swing and the tree. 
We disentangle the benefits of additional sensor modalities from our model 
contributions, highlighting the advantage of 
gated convolutions 
and our DF-based approach towards reducing blur. 

\noindent\textbf{Quantitative results} are reported in Tab.~\ref{tab:arch_results}. The smallest architecture \textbf{ResNet18}~\cite{godard2019digging} performs worst. The larger U-Net \textbf{ResNet50} performs better, and has been generally adopted~\cite{blanchon2020p2d,monodepth17}. Note \textbf{p2d}~\cite{blanchon2020p2d} uses a different data representation (Stokes) for polarisation \cf \textbf{ResNet50}; 
performance decreases. We believe the Stokes representation, using angle directly, is more sensitive to noise and not appropriate for an SSIM loss with the self-supervised stereo \textbf{(S)} training strategy.
MiDaS~\cite{Ranftl2020} provides second best performance 
and yet necessitates roughly 
${\times}2$ GMACs. Our architecture provides best performance while remaining relatively compact which we largely 
attribute to 
gated convolutions 
and displacement field estimation (see Sec.~\ref{sec:arch}).
\vspace{-4pt}
\section{Conclusion}
\vspace{-4pt}
\label{sec:conclusion}

We systematically investigate the effect of using 
additional information from co-modal sensors at training time, for the task of 
monocular depth estimation from polarisation imagery. Our exploration is enabled through a unique multi-modal video dataset 
which constitutes 
synchronized images from binocular polarisation, raw i-ToF and structured-light depth. 
We quantify 
the beneficial influence of both \emph{passive} and \emph{active} sensors, leveraging self-supervised and cross-modal 
learning strategies that lead to the proposal of a new method providing sharper and more accurate depth estimation. 
This is made possible through two physical models that describe the relationships between polarisation and surface normals on one side 
and correlation measures and scene depth on the other.
We 
believe that our fundamental investigation of modality combination and the CroMo dataset can accelerate research of both spatial and temporal fusion, towards advancing cross-modal computer vision. 



{\small
\bibliographystyle{ieee_fullname}

}

\clearpage
\section{CroMo: Supplementary Material}
\label{sec:supplementary}\noindent


\setcounter{equation}{0}
\setcounter{figure}{0}
\setcounter{table}{0}
\setcounter{page}{1}
\renewcommand{\theequation}{S\arabic{equation}}
\renewcommand{\thefigure}{S\arabic{figure}}
\renewcommand{\thetable}{S\arabic{table}}

We provide additional materials to supplement our main paper. In Sec.~\ref{sec:supplementary:polarisation} we provide observations on the properties of light polarisation. Sec.~\ref{sec:supplementary:surface_normal} states the specifics for our surface normal estimation process.
In Sec.~\ref{sec:supplementary:bundle}, we provide additional details for our multi-view camera calibration procedure, Sec.~\ref{sec:supplementary:model_details} provides some further modelling details and finally Sec.~\ref{sec:supplementary:arch} gives supplementary information on our network architectures and learning parameters.

\subsection{Light polarisation parameters}
\label{sec:supplementary:polarisation}

\begin{figure}[h!]
    \centering
    \begin{subfigure}[t]{0.4925\textwidth}
        \centering
        \includegraphics[page=1, trim = 1380 1850 520 390, clip,width=0.95\linewidth]{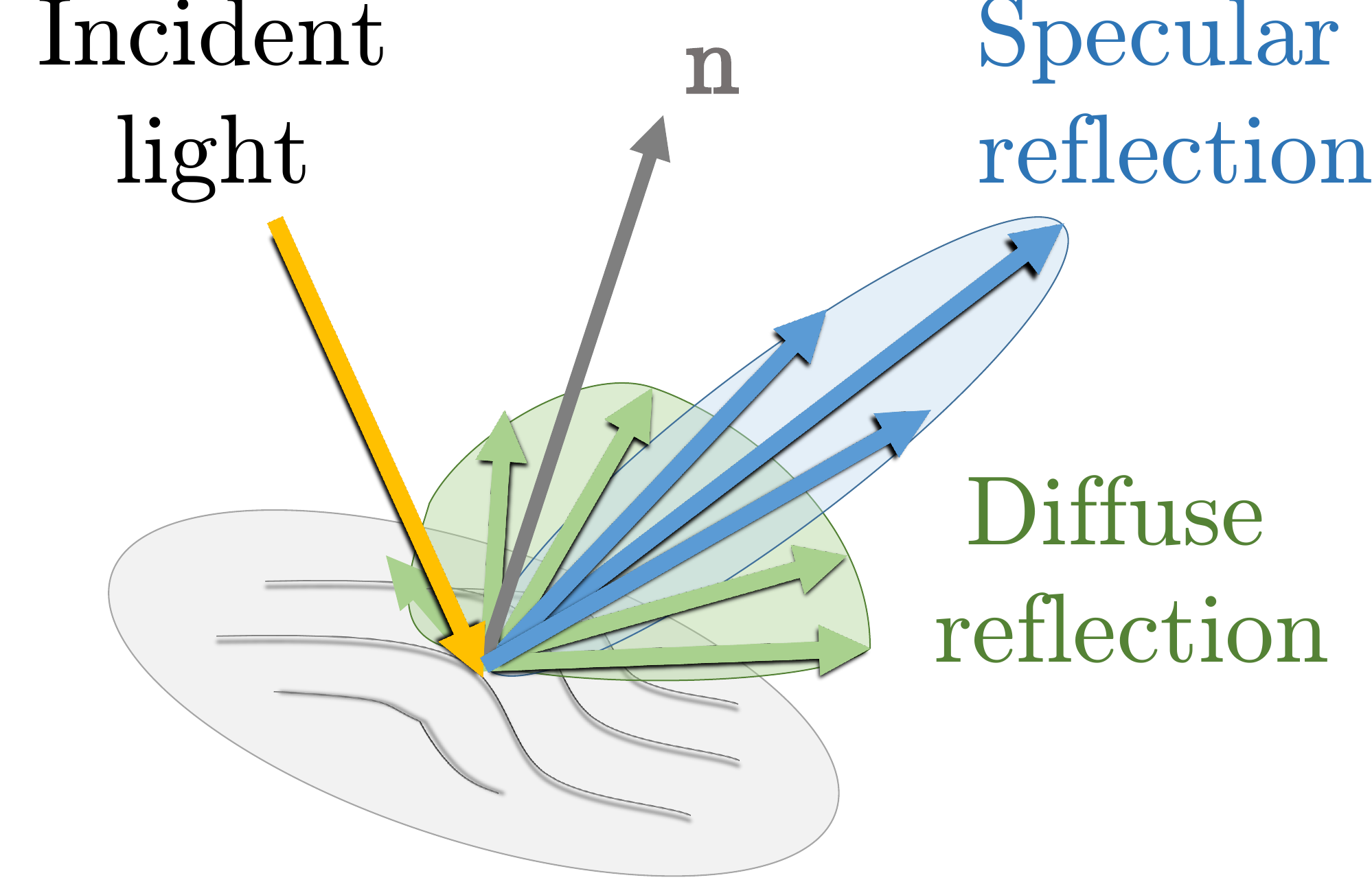}
        \caption{}
    \end{subfigure}%
    ~ 
    \begin{subfigure}[t]{0.4925\textwidth}
        \centering
        \includegraphics[page=1, trim = 300 1850 1600 390, clip,width=0.95\linewidth]{polarisation.pdf}
        \caption{}
    \end{subfigure}
    \caption{(a) differing types of reflected light and (b) the link between a surface normal \textbf{n}, its viewing angle $\theta$ and its azimuth angle $\alpha$ (right).
 }
\label{fig:light reflection and polarisation geometry}
\end{figure}



Most natural light sources emit unpolarized light that only becomes polarized if reflected. Hence the type of reflection, illustrated in Fig.~\ref{fig:light reflection and polarisation geometry}, either \textit{diffuse} ($d$) or \textit{specular} ($s$), influences the characteristics of the reflected polarized light. More specifically, the reflective surface influences the relation between the normals' parameters $(\theta,\ \alpha)$ and the polarisation parameters $(\rho, \ \phi)$,  defined in
Eq.~\ref{eq:rho} and \ref{eq:Polarisation phase} of the main paper.

\subsection{Surface normals}
\label{sec:supplementary:surface_normal}

In 
the main manuscript we estimate polarisation intensity using the varying coordinates of surface normals. Hence, the computation of these normals, derived from network depth prediction, plays an important role in the  training process. To increase the robustness of estimated normals, we compute the cross products using four distinct pairs of orthogonal directions as in~\citeS{yang2018unsupervised}: 

\begin{equation}
    \left\{
    \begin{array}{ll}
        \textbf{n}_{0} &= 
        \partial_{x}\textbf{v} \times \partial_{y}\textbf{v}\\
        
        \textbf{n}_{1} &= \partial_{-x}\textbf{v} \times \partial_{-y}\textbf{v}\\
        
        \textbf{n}_{2} &= \partial_{x+y}\textbf{v} \times \partial_{-x+y}\textbf{v}\\
        
        \textbf{n}_{3} &= \partial_{-x-y}\textbf{v} \times \partial_{x-y}\textbf{v}
    \end{array}
    \right.
\label{eq:Normal computation}
\end{equation}
       
\noindent The weighted average of these normals is calculated using weights $w_{i}$ where:
\begin{equation}
    \left\{
    \begin{array}{ll}
        {w}_{0} &= \exp(-0.5\|\partial_{x}i_{un}\|_1) \cdot \exp(-0.5\|\partial_{y}i_{un}\|_1)\\
        
        {w}_{1} &= \exp(-0.5\|\partial_{-x}i_{un}\|_1) \cdot \exp(-0.5\|\partial_{-y}i_{un}\|_1)\\
        
        {w}_{2} &= \exp(-0.5\|\partial_{x+y}i_{un}\|_1) \cdot \exp(-0.5\|\partial_{-x+y}i_{un})\|_1)\\
        
        {w}_{3} &= \exp(-0.5\|\partial_{-x-y}i_{un}\|_1) \cdot \exp(-0.5\|\partial_{x-y}i_{un})\|_1)
    \end{array}
    \right.
\label{eq:normal_weights1}
\end{equation}
\noindent The final surface normal (unnormalized) is then estimated by their linear combination:
\begin{equation}
    \textbf{n} = \frac{1}{4} \sum_{i=0}^{3} w_{i} \cdot \textbf{n}_{i}
    \label{eq:normal_weights2}
\end{equation}
\noindent Weights $w_{i}$ result in neighbouring pixels of $i_{un}$ 
that contain strong color disparity, to be down-weighted in the normal computation. This follows from the assumption that such pixels are more likely to represent different objects. Conversely, if neighbouring pixels possess similar color, they are more likely to correspond to the same object and their associated partial derivatives are more likely to provide normals that accurately describe the observed object shape.

\subsection{Graph-based bundle adjustment}
\label{sec:supplementary:bundle}
As discussed in Sec.~\ref{sec:dataset:capture_rig} of our main paper the calibration of extrinsics, intrinsics and distortion coefficients, for all four capture-rig cameras, is achieved using a graph-based bundle-adjustment~\citeS{g2o} that improves multi-view calibration. We provide here further details of our multi-view calibration procedure. 

We start with well established calibration methods~\citeS{heikkila1997four}
to obtain the intrinsics $K_k$ and distortion coefficients $d_k$ for each camera $C_k$, where $k \in \{0,1,2,3\}$. We use a standard pinhole camera model and define $C_0$ as the left polarisation camera, $C_1$ the right polarisation camera, $C_2$ the i-ToF camera, and $C_3$ the structure light camera. We use five parameters for the distortion coefficients and collect $n$ images of a calibration checkerboard, from all cameras synchronously. In practice we move the checkerboard in front of the cameras while keeping the camera rig stationary. We attempt to cover as wide a field-of-view as possible for all four cameras. We find it is more important to thoroughly cover and account for the extremities of the individual images as opposed to attempting to be visible to all cameras simultaneously. Further, we estimate the rigid transformation for each camera pair composed of $C_0$ (our world reference), and camera $C_k$ in turn, where $k \in \{1,2,3\}$. This provides the extrinsics $T_{k \rightarrow 0} = [R_{k \rightarrow 0} \big| t_{k \rightarrow 0}]$ for camera $C_k$ (with $T_{0 \rightarrow 0} = [I \big| 0]$).\\

These initial intrinsic, extrinsic parameter values and the distortion coefficients are however sub-optimal as they are obtained by solving successive sub-optimisation problems. Towards improving the multi-camera calibration, we define the reprojection error of points $X^j$ on the image $I_i$ for the camera $C_k$ as
\begin{equation}
\begin{aligned}
\widehat{x_{j}^{i}} &= \pi\left( T_{k \rightarrow 0}, T_0^i, X^j,d_k,K_k\right) \\
E_k^i &=
\sum_{j=0}^{\text{\#points}} 
 \mathbb{1}_{\widehat{x_{j}^{i}} \in I_i} \cdot 
 \text{dist}\left(x_{j}^{i}, \widehat{x_{j}^{i}} \right)^{2}
 \end{aligned}
\label{eq:ba_error}
\end{equation}
\noindent Where $T_0^i$ is the position of camera $C_0$ for image $i$, and $\widehat{x_{j}^{i}}$ is the distorted 2D point from the projection function $\pi(\cdot)$ which projects a 3D Point $X^j$ visible by the camera $C_k$ at position $T_0^i \cdot T_{k \rightarrow 0}$ with distortion coefficients $d_k$ and intrinsic parameters $K_k$ on  image $I_i$.
The function $\text{dist}(\cdot)$ defines the robustified distance between 2D points, \ie a Huber $m$-estimator, and $x_{j}^{i}$ is the 2D point detected on the checkerboard with a corner detector corresponding to the 3D point $X^j$ in image $I_i$.
The indicator function $\mathbb{1}_{\widehat{x_{j}^{i}} \in I_i}$ defines whether the 2D Point $\widehat{x_{j}^{i}}$ is visible in image $I_i$.\\

\noindent Finally, we used a graph-based bundle-adjustment~\citeS{g2o} to model the global problem, for all cameras $C_k$, jointly as:
\begin{equation}
\min _{T_0^i, T_{k \rightarrow 0}, d_k, K_k }
\sum_{k=0}^{\text{\#cameras}}
\sum_{i=0}^{\text{\#images}} 
 E_k^i,
\label{eq:ba_min}
\end{equation}
\noindent with $T_0^0$ fixed to $[I \big| 0]$ in order to properly constrain the gauge freedom.
All camera calibration parameters are initialised using the values obtained from the original individual calibrations.\\

This formalism, borrowed from the SLAM community~\citeS{triggs1999bundle}, allows us to optimize all parameters, \ie the intrinsic, the extrinsic and the distortion parameters for all cameras, jointly. We find the global optimisation process is able to improve our calibration RMSE by ${\sim}5$\textendash$10\%$.

\subsection{Additional modelling details }
\label{sec:supplementary:model_details}

\subsubsection{Reflection ambiguities}
\label{sec:supplementary:model_details:reflections}

A diffuse-specular ambiguity initially exists in our formulation; pertaining to diffuse or specular reflection (see Eq.3, main paper). This ambiguity is addressed during training via the $\min$ operator found in Eq.18. We propose to resolve reflection ambiguities (per-pixel) by minimization of the SSIM loss between respective \{specular, diffuse\} images and the input image, towards consistently providing a valid training signal. Secondly, the azimuthal $\pi$-ambiguity is directly accounted for by the formulation of Eq.1; the inherent $\cos(\cdot)$ modulation nullifies ambiguity found in its input ($2\phi$ component of the argument) and thus supervision is not adversely affected due to $\phi$ being modulo $\pi$.



\subsubsection{Wrappings ambiguities}
\label{sec:supplementary:model_details:wrappings}

An analytical solution exists for the correlation to depth transform and a wrapping ambiguity remains. However we highlight that a reconstructed depth, although ``phase wrapped'', is still able to provide reliable surface normals that can be used to produce (1) the degree of linear polarisation and (2) the Angle of Polarisation, for both diffuse and specular surfaces. 
Once these are projected to the $N2$ referential, this information is used in conjunction with the brightness of the left polarisation image to render valid ``{Recovered polarisation}'' images, (see Fig~\ref{fig:loss} of our main paper). 

\subsubsection{Polarization intensity recovery}
\label{sec:supplementary:model_details:pol_recovery}

To render the intensity, we require the brightness of each pixel ($i_{un}$ in Eq.1). We obtain the brightness of the polarisation image by channel-wise summing of the left polarisation input pixel values. Two images are rendered following Eq.1; for both the cases of diffuse and specular images. We use a binary mask to select values, \emph{pixel-wise}, from either the specular or diffuse image. The mask selects pixels such that the minimum SSIM loss between the \{specular, diffuse\} image and the input image are retained. 

The two images formed therefore constitute only an \emph{intermediary} step towards producing a final image. We use a binary mask to then select values, \emph{pixel-wise}, from either the specular or diffuse image to form a new image containing the pixels that retain the minimum SSIM loss between the \{specular, diffuse\} image and the input image (\ie the $\min$ in Eq.18 is \emph{per pixel}). We thus form a final image that contains both specular and diffuse components. \\

\subsubsection{Correlation image rendering from depth}
\label{sec:supplementary:model_details:corr_from_depth}

Analogous to the Polarization image strategy, we use the input correlation image (obtaining $\alpha$, $\beta$ estimates), in addition to depth information, to estimate both the ambient $\beta$ and reflectance $\alpha$, for correlation reconstruction.

\subsection{Architecture and training details}
\label{sec:supplementary:arch}


\subsubsection{Architecture}

We provide additional description for the network architecture that we propose in order to process the considered input modalities. Instances of this architecture are depicted as edges `$N1$' and `$N2$' in the system design; see Fig~\ref{fig:model} of our main paper.\\


We employ a standard U-Net architecture, similar to our baseline~\cite{godard2019digging}, including skip connections. The encoder is based on a `Resnet'~\cite{he:2016} style block, with the original convolutional layer replaced by gated convolution \cite{yu2019freeform}. The size of the input images are 
$512{\times}544{\times}12$ for polarisation and $640{\times}480{\times}4$ for i-ToF, respectively. 

For polarisation, we have the following configuration; layer one: $512{\times}544{\times}64$, layer two: $256{\times}272{\times}128$, layer tree: $128{\times}136{\times}256,$ layer four: $64{\times}68{\times}512$.
For i-ToF, we have the following configuration; layer one: $640{\times}480{\times}64$, layer two: $320{\times}240{\times}128$, layer tree: $160{\times}120{\times}256,$ layer four: $80{\times}60{\times}512$. 
Both depth and Displacement Field decoders are a standard cascade of convolutions with layer resizing. Encoder skip connections are concatenated after each resize operation (see Fig.~\ref{fig:model}).\\ 

\subsubsection{Training parameters}
To aid reproducibility, we report training parameters and hyperparameters. We use identical training parameters and align with our baseline~\cite{godard2019digging} where possible. We use the Ranger optimiser \citeS{yong2020gradient} and batch sizes of 
$8$, 
a learning rate of $1e{-}4$ with an exponential learning rate decay. We train all considered methods for $50$ epochs.

\subsubsection{Comparison with RGB input} 
A direct comparison with RGB input forms a relevant and interesting line of enquiry. Our custom capture rig does not currently accommodate this modality directly. However, towards investigating this experimentally, we did transform the polarisation input frame to an RGB frame by considering the polarisation intensity of each RGB channel, individually.
We note that this is \emph{not} directly equivalent to an RGB sensor since the bayer pattern differs.
We actively decided \emph{not} to include this experimental work in the main paper to avoid misinterpretation and confusion. 
Preliminary work evaluating our Polarisation input \cf the noted ``\emph{Polarisation-converted-to-RGB}'' 
showed improvements using Polarisation (RMSE 1.4) over ``\emph{Polarisation-converted-to-RGB}'' (RMSE 1.53). 

\subsubsection{Controlling for capture environment} We note that the i-ToF modality excels in indoor environments, however these represent a relatively smaller portion of our dataset. To corroborate this, we report an experiment that considers our various training strategies, tested on only an indoor environment (\emph{Kitchen}). The addition of i-ToF (from (\textbf{S}) to (\textbf{ST})), at training time, significantly improves the predicted depth in this restricted setting (see Tab.~\ref{tab:kitchen}).

\begin{table}[h]
\centering
\begin{adjustbox}{width=0.8\textwidth}

\begin{tabular}{ c | c | c c c } 
\toprule
{ \small{\shortstack{Image\\sensors}}} &
{ \small{\shortstack{Training\\strategy}}} &
{\centering Sq Rel}  &
{\centering {RMSE}} & 
{\centering RMSE Log} 

\tabularnewline
\midrule
2 & \small{(\textbf{S})}    & 0.6202 & 1.2930 & 0.2944 \\
3 & \small{(\textbf{ST})}   & 0.3001 & 0.6520 & 0.2237 \\
4 & \small{(\textbf{STLM})} & \textbf{0.2105} & \textbf{0.5431} & \textbf{0.180} \\ 
\bottomrule
\end{tabular}
\end{adjustbox}
\vspace{0.5em}
\caption{Test on \emph{Kitchen} scene (780 frames): 
additional sensors can be observed to improve performance. The largest improvement comes from the addition of the i-ToF, (from (\textbf{S}) to (\textbf{ST})), in an exclusively indoor test setting.}
\label{tab:kitchen}
\end{table}


\subsubsection{Further analysis of \emph{where} additional sensors help}

\begin{figure}[h!]
    \centering
    \begin{subfigure}[t]{0.2425\textwidth}
        \centering
        \includegraphics[trim = {300 136 90 200}, clip, width=0.95\linewidth]{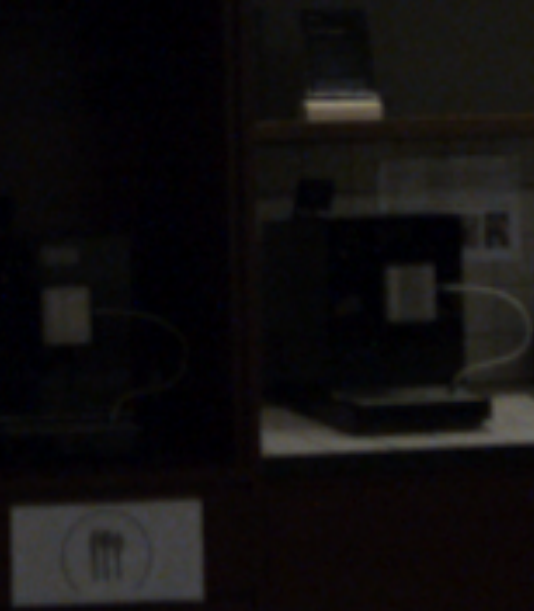}
        \caption{Pol. Intensity}
    \end{subfigure}%
    ~
    \begin{subfigure}[t]{0.2425\textwidth}
        \centering
        \includegraphics[trim = {300 136 90 200}, clip, width=0.95\linewidth]{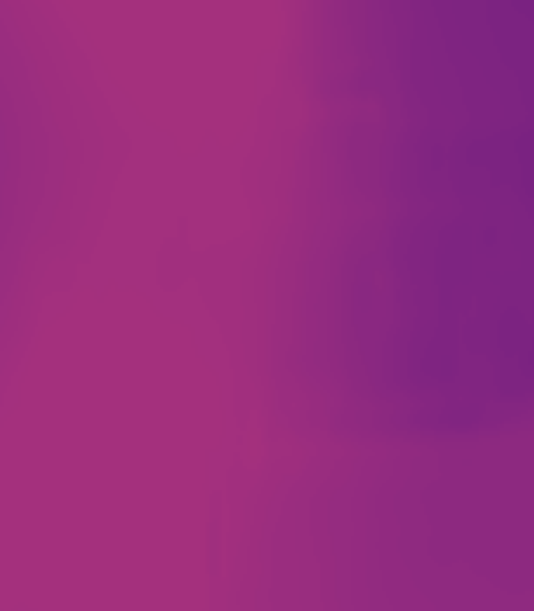}
        \caption{\textbf{S}}
    \end{subfigure}%
    ~
    \begin{subfigure}[t]{0.2425\textwidth}
        \centering
        \includegraphics[trim = {300 136 90 200}, clip, width=0.95\linewidth]{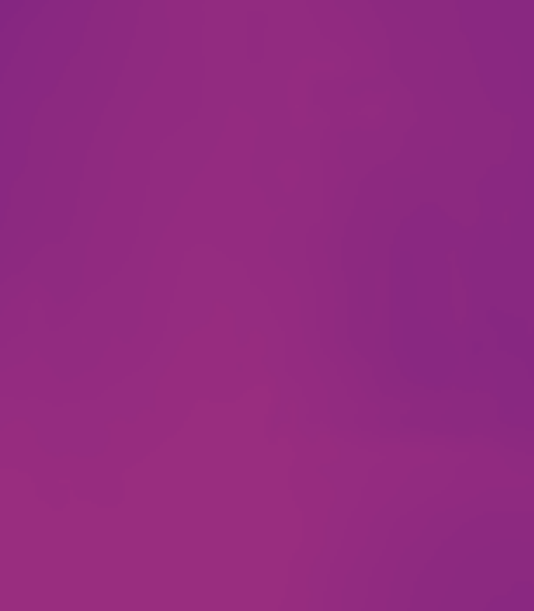}
        \caption{\textbf{ST}}
    \end{subfigure}%
    ~
    \begin{subfigure}[t]{0.2425\textwidth}
        \centering
        \includegraphics[trim = {300 136 90 200}, clip, width=0.95\linewidth]{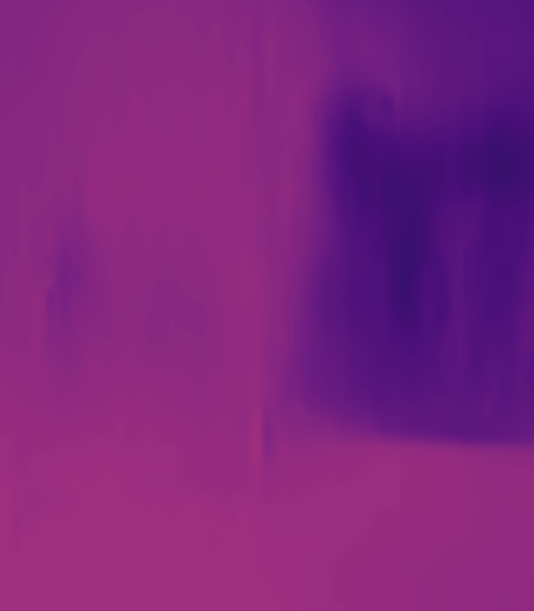}
        \caption{\textbf{STLM}}
    \end{subfigure}
    \caption{Depth estimation improvements possible from a common input (a). We show gains achievable by adding orthogonal signal during training, where inference utilises only a single polarisation image in all cases. See text for further detail.}
\label{fig:where_helping}
\end{figure}

We include preliminary further analysis with respect to investigation of scenarios {where} additional sensors help. We include an example that highlights two points (see Fig.~\ref{fig:where_helping}). Due to the concave nature of the scene, the addition of ToF information alone during training (from \textbf{S} to \textbf{ST}) adversely impacts the depth prediction and we find MPI often detrimental to the ToF sensor in such cases. Additional sensors (from \textbf{ST} to \textbf{STLM}) do however improve final depth estimation and we show gains achievable by adding orthogonal signal during training, where inference utilises only a single polarisation image in all cases. Additional investigation and rigorous analysis of such scenarios makes for interesting future work.  

\subsubsection{Additional structured light experiments}
\label{supp::rs_only} 
The structured light sensor present in our camera rig offers low-noise signal which we find can also be leveraged in a supervised fashion, directly. For completeness, we compare the resulting depth estimation when supervising directly with structure-light (\textbf{D}) and our approach, using unsupervised signals (\textbf{STLM}). The structure-light signal, obtained from our Realsense sensor, is claimed reliable up to a 10 meters range according to the constructor~\cite{realsense}. We thus further investigate by evaluating performance over distinct $0-10m$ and $0-20m$ ranges. Results in Tab.~\ref{tab:rsonly} show that the fully supervised method (\textbf{D}) can offer similar performance to our approach (\textbf{STLM}) in the range $0-10m$ yet performance degrades by significant margins when considering the more challenging $0-20m$ range. This highlights the benefits of our unsupervised multi-modal strategy (\textbf{STLM}); leveraging information from multiple sensor sources and an ability to learn to adapt when a particular sensor results in low quality measurement, due to unsuitable physical conditions (\eg structured light in the $10-20m$ range). 
\begin{table}[h]
\centering
\begin{adjustbox}{width=0.95\textwidth}
\begin{tabular}{ c | c | c c c | c c c } 
\toprule
\multicolumn{1}{c|}{\multirow{2}{*}{\small{\shortstack{Image\\sensors}}}} & \multirow{2}{*}{ \small{\shortstack{Training\\strategy}}} & \multicolumn{3}{c|}{0-10m} & \multicolumn{3}{c}{0-20m}\\
\multicolumn{1}{c|}{}                  &                   & 
{\centering Sq Rel}  &
{\centering {RMSE}} & 
{\centering RMSE Log} &
{\centering Sq Rel}  &
{\centering {RMSE}} & 
{\centering RMSE Log}   
\tabularnewline
\midrule
1 & \small{(\textbf{D})}  &\textbf{0.9479}	&\textbf{1.4246}	&\textbf{0.2117}	&5.447	&6.2629	&1.6134\\
4 & \small{(\textbf{STLM})}

&1.0031&1.4889& 0.2527 &\textbf{1.3994} &\textbf{2.9512}	&\textbf{0.3879} \\ 

\bottomrule
\end{tabular}
\end{adjustbox}
\vspace{0.5em}
\caption{Comparison of training strategies for two depth prediction ranges. Our training strategy (\textbf{STLM}) works well in spite of the operational limits of particular sensors. }
\label{tab:rsonly}
\end{table}

\subsubsection{Additional details on the $\mathcal{L}_{\text{struct}}$ loss}
We select to use a structured light loss similar to the loss proposed in~\cite{watson-2019-depth-hints}. We find that such indirect supervision of the structure light signal allows to automatically select the best source of information, particularly in situations where the structure light signal fails or becomes unreliable (as discussed in Sec.~\ref{supp::rs_only}). Formally, given a depth from the structured light $D_{\text{struct}}$, the loss reads:
\begin{align}
 \mathcal{L}_{\text{struct}} &= E_{\text{pe}}\left(I_{l}, I_{\text{right}\underset{D_\text{struct}}{\longrightarrow} \text{left}}\right)
\end{align}

We make use of an additional $\mathcal{L}_1$ loss between predicted depth and the $D_{\text{struct}}$ depth, when $\mathcal{L}_{\text{struct}}$ is minimal (see ~Eq.\ref{eq::total_loss} in the main manuscript).

\subsubsection{Limitations and Societal Impact}
\noindent\textbf{Limitations} We note distinct limitations that relate to our sensor setup. Active sensors have limited range and areas of operativity \eg i-ToF often offers weaker performance outdoors, structure light sensors are of limited range, and polarisation sensors sacrifice spacial sampling resolution for spectral sampling resolution. Our multi-modal ideas attempt to combat these limitations indirectly however we remain bound by the physical laws of light. 

Additionally, our current hardware setup is operable by a single person, and yet data capture is currently more cumbersome than \eg use of a modern smartphone. Training data collection, that involves the acquisition of multiple modalities, currently induces a somewhat larger investment of effort over monomodal capture. With the argument being that the cost may then be recouped when assessing monocular inference time performance. Our hardware rig constitutes a research prototype and form factor likely improves as camera evolution results in further reductions in sensor size, weight and cost.

Finally we would note that our current dataset does not yet capture all possible scenarios and represents but a subset of urban scenes where depth estimation can prove valuable. Future capture sessions will look to enrich and widen the recorded capture scenarios, towards increasing the value of the data resource that we provide to the community. \\

\noindent\textbf{Societal Impact}
We note that our proposed CroMo dataset was collected by only two human operators 
in urban environments. While care was taken towards objective scene capture, such collected data may yet reflect the biases of human operators; influencing specific content, scenarios or capture setups. Efforts towards the reduction 
of bias, introduced by manual human operators, might suggest mounting of the system on automatic vehicles in future.  
Additional ideas, toward mitigation of the axis of bias relating to manual data capture, can be considered an interesting future research direction.
{\small
\bibliographystyleS{ieee_fullname}
\bibliographyS{S.bbl}
}

\end{document}